\documentclass{article}

\usepackage{arxiv}

\usepackage[utf8]{inputenc} % allow utf-8 input
\usepackage[T1]{fontenc}    % use 8-bit T1 fonts     
\usepackage{url}            % simple URL typesetting
\usepackage{booktabs}       % professional-quality tables
\usepackage{amsfonts}       % blackboard math symbols
\usepackage{nicefrac}       % compact symbols for 1/2, etc.
\usepackage{microtype}      % microtypography
\usepackage{lipsum}
\usepackage{multirow}
\usepackage{amsmath}
\usepackage{graphicx}
\graphicspath{ {./images/} }
\usepackage[authoryear, round]{natbib}
\usepackage{float}
\restylefloat{table}
\usepackage{threeparttable}
\usepackage[table]{xcolor}
\definecolor{Gray}{gray}{0.3}
\usepackage{array}
\newcolumntype{C}[1]{>{\PreserveBackslash\centering}p{#1}}
\newcolumntype{R}[1]{>{\PreserveBackslash\raggedleft}p{#1}}
\newcolumntype{L}[1]{>{\PreserveBackslash\raggedright}p{#1}}

\title{CORTEX: A Cost-Sensitive Rule and Tree Extraction Method}

\author{
Marija Kopanja \\
  Department of Mathematics and Informatics, \\ Faculty of Sciences, \\ 
  University of Novi Sad\\
  Trg Dositeja Obradovića 3\\
  Novi Sad, 21000, Serbia \\
  \texttt{marija.kopanja17@gmail.com} \\
   \And
 Miloš Savić\\
  Department of Mathematics and Informatics,\\ Faculty of Sciences, \\ 
  University of Novi Sad\\
  Trg Dositeja Obradovića 3\\
  Novi Sad, 21000, Serbia \\
  \texttt{svc@dmi.uns.ac.rs} \\
  \And
 Luca Longo\\
  Artificial Intelligence and Cognitive Load Research Lab, \\ The Centre of Explainable Artificial Intelligence, \\ Technological University Dublin \\
  Central Quad, CQ-214 Grangegorman Campus\\
  Dublin, D07 ADY7, Ireland \\
  \texttt{luca.longo@tudublin.ie} \\
}

\begin{document}
\maketitle
\begin{abstract}
Tree-based and rule-based machine learning models play pivotal roles in explainable artificial intelligence (XAI) due to their unique ability to provide explanations in the form of tree or rule sets that are easily understandable and interpretable, making them essential for applications in which trust in model decisions is necessary. These transparent models are typically used in surrogate modeling, a post-hoc XAI approach for explaining the logic of black-box models, enabling users to comprehend and trust complex predictive systems while maintaining competitive performance. Explaining complex models can be crucial in class imbalance frameworks due to potential biases in the model being explained. Potential bias could lead to misleading interpretations and unfairness in the decision-making process. The cost-sensitive decision tree (CSDT) method belongs to the cost-sensitive learning methods that can be used in the imbalanced learning framework by considering a class-dependent cost matrix, a matrix associated with classes instead of individual samples. This study proposes Cost-Sensitive Rule and Tree Extraction (CORTEX) method, a novel rule-based XAI algorithm grounded in the multi-class CSDT method. The original version of the CSDT method is introduced only for the binary classification problems. Here, the framework is extended to classification problems with more than two classes by inducing the concept of an n-dimensional class-dependent cost matrix into the CSDT algorithm. The CORTEX provides a visual explanation through a decision tree model, a hierarchical structure composed of nodes and branches. The rules from CORTEX are created in conjunction with feature test conditions (antecedents), while the rule consequence is obtained as a class label held by the terminal node. The performance of CORTEX as a rule-extractor XAI method is compared to other post-hoc rule extraction methods across several datasets with different numbers of classes. Several quantitative evaluation metrics are employed to assess the explainability of generated rule sets. Our findings demonstrate that CORTEX, as a rule-based model-agnostic XAI method, is competitive with other tree-based methods and can be superior to other rule-based methods across different datasets. The extracted rule sets suggest the advantages of using the CORTEX method over other methods by producing smaller rule sets with shorter rules on average across datasets with a diverse number of classes. Overall, the results underscore the potential of CORTEX as a powerful XAI tool for scenarios that require the generation of clear, human-understandable rules while maintaining good predictive performance.
\end{abstract}

% keywords can be removed

\section{Introduction}
Explainable artificial intelligence (XAI) is one of the fastest emerging sub-fields of AI dedicated to developing methods for making machine learning (ML) models more understandable and transparent. When data is highly non-linear and complex, sophisticated neural network models must be trained to gain knowledge and insights. For the majority of real-world application problems, the models must be as accurate as possible, regardless of their interpretability. However, the inference process of black-box models is incomprehensible to humans. Consequently, XAI techniques play a crucial role in making the inference process of these models more understandable. In application domains where comprehensibility and generalization ability are of equal importance, there is a motivation to combine the generalization ability of deep neural network models with the explainability given by XAI techniques. With an aim to extract information from already trained models, several different methods are developed for explaining their inferential process post-hoc, which means after the model has been trained. Post-hoc XAI methods do not modify the internal structure or training process of the original model being explained. Instead, they provide retrospective insights into the model's decisions.
Existing XAI methods can be categorized in various ways, depending on the input data type, the machine learning model being explained, and the scope of the explanations, among others. Several studies \citep{manifesto, ali, Vilone2021, luca_review} overview the concepts of XAI, current research and trends, and different taxonomies. For example, in \citep{Vilone2021} one of the division dimensions is the stage at which a method generates explanations. Ante-hoc XAI methods are considered naturally understandable, while post-hoc XAI methods mimic the behavior of an underlying ML model by using an external explainer.  Accordingly, the CORTEX method proposed in this study is post-hoc model-agnostic, meaning it can be applied and used to explain any pre-trained ML model. In \citet{ali} is a proposed taxonomy for post-hoc explainable methods, grouping them into six families, among which are attribution methods, which assume evaluating the relative significance of input features in decisions of a given ML model. Creating a surrogate model (imitation of the original model) is a post-hoc approach \citep{ali} used to clarify the AI model’s decisions by approximating the original model. In line with the taxonomy, creating a surrogate model is one use of the perturbation approach, a sub-group of the attribution methods family. The proposed CORTEX can be considered a tree-based and rule-based surrogate method for any ML model, including dense neural networks. Developing a tree-based/rule-based model as an explanation for a neural network model constitutes a pedagogical or decompositional approach; the former method induces a tree model on a relabeled target variable without using internal elements of the network, in contrast with the decompositional approach in which focus is on extraction of rules separately form each unit of the neural network \citep{Schmitz1999}.

The proposed CORTEX builds upon the cost-sensitive decision tree (CSDT) method. As in any tree-based ML algorithm, the original CSDT proposed by \citet{Correa_CSDT} \footnote{In the source code of the CSDT, there is no implementation for the visualization of the obtained tree model. The code is available in the \textit{costcla} repository \url{https://pypi.org/project/costcla/}} could be represented as a flowchart-like structure. That is, the method could provide a visual explanation through a decision tree model, a hierarchical structure composed of nodes and branches. Besides, every path from the root node of a tree model to any of the terminal nodes can be converted to a rule by gathering all the feature test conditions appearing in the path. In this way, the rules from CORTEX are created as a conjunction of antecedents, while the rule consequence is obtained as a class label held by the terminal node. The motivation for transforming an inherently tree-based explanation model generated from CORTEX into a set of rules arises from the observation that analyzing rules instead of following paths from the root node to a terminal node often improves comprehensibility \citep{freitas}.  Furthermore, the ability of original CSDT to generate shallower decision trees than traditional decision tree algorithms is observed in the study by \citet{kopanja} and represents a motivation for developing CORTEX. However, as highlighted and explored in our previous paper \citep{kopanja}, the CORTEX is only one part of a multi-faceted idea of a novel model-agnostic XAI framework \citep{kopanja2024}. The transformation from a tree model into a set of rules is also essential to facilitate comparisons with other tree-based and rule-based XAI methods. 

The experimental section compares the proposed CORTEX method to widely used XAI rule-extraction algorithms, including C4.5-PANE, REFNE, RxREN, and TREPAN. Once considered state-of-the-art in the domain, these methods are still frequently used as benchmarks in similar studies \citep{giulia_2020, giulia_2021}. They are also inherently diverse, including tree-based and pure rule-based methods. The importance of comparing different XAI methods is emphasized in other studies \cite{ribeiro2016}. The selected subset of benchmarking methods is extended with a traditional decision tree classifier, induced using Classification And Regression Trees - CART algorithm \citep{CART}, to provide a more well-rounded evaluation.

The existing rule-extractors proposed in the literature have certain limitations — such as reduced interpretability due to many rules and long rule lengths on average — which our CORTEX method aims to address without decreasing other general measures related to the XAI method's predictive capacity. By comparing these methods, we strive to demonstrate that our CORTEX offers competitive performance while addressing key limitations such as a minimized number of rules and average rule length. Extracting a set of rules from each method facilitates comparisons among them; these rules replicate the inferential process of the two underlying neural network models trained on eight datasets. 

After generating the rule sets, we assess the rules' degree(s) of explainability. Assessment of explanations is one of the four key aspects that one needs to consider when employing the XAI method \citep{ali}. An objective, quantitative assessment of explanations includes a set of computational metrics for evaluating the validity of the produced rules. Completeness is used to evaluate if all samples are covered by the given set of rules. The fidelity measure reflects how closely the explanations match the prediction model’s behavior. The accuracy of a rule-extractor is expressed as a degree of correctness and refers to the generalization of an explanation of a specific decision to previously unknown situations. Another desirable property of explanations is reliability, which ensures that minor input modifications do not significantly influence the XAI model prediction - expressed as robustness in our evaluation process. The comprehension of obtained rule sets refers to the readability and length of explanations; they are evaluated by counting the number of rules and the average number of antecedents across rules (average rule length). 

The remainder of this manuscript is organized into five sections. The next section provides an overview of post-hoc XAI approaches used to generate explanations for ML models, focusing on the rule extractors and tree-based models. Section 3 presents the CORTEX method after briefly introducing the CSDT method and the concept of n-dimensional cost matrix. Section 4 describes the experimental setup for evaluating the proposed method and metrics employed to assess and compare the degree of explainability of generated sets of rules. Further, Section 5 gives a CORTEX comparison with other methods and discusses the findings of the reported results. Finally, in Section 6, the main contribution of this paper is summarized, and directions for future work are defined.

\section{Related work}
In many use cases, the model of choice for a given application will be a complex neural network model due to the high-performance capacity of these models. Nevertheless, interpretability has also been emphasized in recent years. While models with high accuracy need additional explainability, there are models with low accuracy that are still simpler to comprehend. The interpretability-accuracy trade-off is a well-known concept (\cite{Rivera2024, ribeiro2016, darpa, RizzoL18Explainability}): the more accurate the model, the less it is understandable. Tree-based models are considered \cite{ali} self-interpretable and comprehensible; they explain the decisions and logically show them in a hierarchical structure illustrated as a directed graph. However, oversimplified explanation models such as shallow decision trees might not be acceptable for some stakeholders \citep{freitas}. A single-tree model is reasonably interpretable. However, an ensemble of tree models such as Random Forest \citep{RF} or other tree-ensemble models, including a cost-sensitive one, is less interpretable but can attain higher accuracy. A limited amount of literature deals with the XAI methods and cost-sensitive machine learning models on tabular data. For example, cost-sensitive CatBoost classifier and LIME explainer are utilized in \citep{Maouche2023} to provide patient-level explanations within computer-aided prognosis systems for breast cancer metastasis prediction. In the study \citep{kopanja2023} is a proposed extension of SHAP method \citep{shap} for tree-based cost-sensitive models, and in their later work \citep{kopanja} the proposed methodology is extended on cost-sensitive ensemble models. 

In the literature, few articles deal solely with interpretable machine-learning models for imbalanced tabular data. Some recent studies \citep{Mustari2023, Dablain2023} have attempted to provide explanations for convolutional neural networks trained on medical images. Having an imbalanced dataset could lead to a biased ML model. Hence, providing explanations might be crucial for such models to ensure fairness, which means the model makes unbiased decisions without favoring any class in the data distribution. The problem of biased inferences is attempted to be tackled through the General Data Protection Regulation (GDPR) that sets out the right to obtain an explanation of the automated decision-making by ML model \citep{vilone_review}. Explaining complex models in class imbalance frameworks due to potential biases in the model being explained could prevent misleading interpretations and unfairness in the decision-making process. Therefore, different XAI tools could enable insights into the biased ML models to improve fairness, transparency, and accountability of the model's decision-making process.

Post-hoc explainability is more critical for domain experts and end-users who are more interested in getting answers on \textit{how} and \textit{why} ML model arrived at a particular prediction and the key features that led to the conclusion. One of the popular post-hoc approaches is creating a surrogate model, the approximation of the underlying black-box model. In the surrogate approach, the aim is to represent the relationship between input data and the output of the neural network model without information on the internal configuration of the network. Surrogate models can be created globally or locally \citep{ali}, where global surrogate models aim to provide an explanation for the model as a whole, and the local surrogate model provides an explanation for a single instance. For example, LIME \citep{lime} and other versions of this approach \citep{alime, tree_alime} are local surrogate models, meaning these methods generate local explanations for individual samples of black-box ML models. Another popular post-hoc method, which can provide both local and global explanations, is the SHAP method proposed by \citet{shap}. The SHAP is based on Shapley values of a conditional expectation function of the ML model. Both model-specific and model-agnostic versions of the SHAP have been proposed. \citet{TreeSHAP} proposed TreeSHAP, a unique model-specific SHAP approximation for tree-based models. AraucanaXAI \citep{Araucana} is post-hoc surrogate approach where classification and regression trees (based on CART algorithm \citep{CART}), are locally-fitted to provide explanations of the prediction of a complex ML model. Comparative evaluation of AraucanaXAI and other local XAI methods, including LIME and SHAP, showed the advantages of AraucanaXAI in terms of high fidelity and ability to deal with non-linear decision boundaries. A combination of decision tree and differential evolution algorithms is proposed in (\citep{Rivera2024}) to develop an evolutionary approach for inducing univariate decision trees that effectively explain the predictions of black-box models. Decision-tree-based surrogate models proved valuable for explaining outliers detected by unsupervised anomaly detection models~\citep{Savic2022}.

Complex deep neural network models can discover complex structures in the data. Still, the learned patterns are hidden knowledge without explicit rules for finding them \cite{ali}. Several approaches have been proposed to explain deep learning classification models, including using decision tree methods as surrogate models and extracting rule sets from the resulting tree \citep{DT_surrogate}. Tree-based algorithm C4.5Rule-PANE \citep{c45} is an extension of a C4.5 decision tree algorithm, capable of extracting if-then rules from ensembles of neural networks, and its performance is compared to other rule-extractors in several studies \citep{giulia_2020, giulia_2021}. Rule Extraction From Neural Network Ensemble (REFNE) was developed to extract symbolic rules from neural networks \citep{refne}. Another rule-based method for enhancing the interpretability of neural networks by translating their complex operations into human-understandable rules is Rule Extraction by Reverse Engineering (RxREN) \citep{rxren}. The RxREN relies on a reverse engineering technique to extract rules from neural networks. Researchers \citep{rxren} have shown that RxREN efficiently prunes insignificant input neurons from the trained neural network models. Rule extraction is a learning problem in the TREPAN \citep{trepan} method that generates a decision tree by querying the underlying network using a query and sampling approach. In \citep{svm_rules}, the performance of TREPAN and C4.5 as rule extractors is assessed using fidelity, correctness, and a number of rules on support vector machine models. The results show that TREPAN obtained the best average performance and consistently outperformed C4.5 with comparable comprehensibility that was more computationally demanding. Another rule-based XAI approach is Anchors \citep{Anchors}, where intuitive and easy-to-comprehend \textit{if-than} rules are generated. 

Decision trees and rule sets are two different representation types of explanations that are easily understandable and interpretable for humans \citep{Guidotti2019}. Nevertheless, both decision trees and rules have their drawbacks \citep{ribeiro2016} related to their graphical and textual representations, respectively. Unlike a decision tree, where a hierarchical structure provides information about feature importance, the importance of a feature is unknown in the textual representation of rules. Users' tolerance for the same explanation type can differ; for example, the number of rules considered too large for some users can be acceptable for others \citep{freitas}. However, some stakeholders might prefer rules as an explanation type instead of the decision tree, and others, depending on their background, can favor decision trees as more comprehensible \citep{ribeiro2016}. Additional experimental studies should be conducted to compare user preferences for various representations of explanations, as none can be considered the best for all applications.

With many XAI methods available, the evaluation process of generated explanations enables practical assessment based on criteria such as transparency, fidelity, robustness, and usability. Different metrics measure different aspects of explanations depending on the type of explanation. For example, suppose explanations are generated in the form of rules. In that case, evaluation metrics such as a number of rules or an average number of antecedents in the rule can be used \citep{giulia_2020}. In the study by \citet{ribeiro2016}, it is argued that average rule length is a fair measure of the simplicity of a rule set. The small input changes should not significantly affect the AI system's behavior. Hence, robustness represents one of the critical measures of the XAI method, where robustness refers to the sensitivity of the AI-driven model output to a change in the input. Beyond these metrics, other quantitative validation factors must be fulfilled by every type of explanation automatically generated by an XAI method \citep{giulia_2021}, including the correctness measured as the portion of samples correctly classified by a given XAI method. 

The effectiveness of explanations generated by some XAI methods must be assessed according to how the explanations aid human users. Therefore, there is an implicit requirement for human-in-the-loop evaluation of AI-driven system reasoning.
Apart from quantitative comprehensive assessment, the AI systems can also be validated by users \citep{ali}. Taxonomy for evaluation of XAI methods is divided into computer-centred and human-centred by \citet{Lopes_2022}, where the former involves methods to obtain a measure of interpretability and fidelity to evaluate the quality of explanations. In contrast, the letter consists of conducting user experiments with human subjects. Human-centred evaluation of explanations in the form of if-than rules is undertaken in the study \citep{Huysmans_2011}. In a more recent study, \citet{dragoni_2020} created a questionnaire to collect feedback from participants on the persuasiveness of automatically generated explanations. Similarly, \citet{giulia_test} have developed and evaluated a novel questionnaire designed to assess the explanations generated by XAI methods reliably. The questionnaire is based on close-ended questions for testing rule-based explanations and tested on argument-based and decision-tree explanations of deep neural networks trained on three datasets over six groups of human participants. Another human-centred study \citep{Anchors} showed that XAI methods could enable users' understanding of the model’s behavior measured by users predicting how a model would behave on unseen samples with less effort and higher precision.

While significant progress has been made within the XAI field in developing new methods and evaluation frameworks, many challenges remain, particularly in existing post-hoc model-agnostic XAI approaches. Surrogate models created by tree-based and rule-based methods can have good predictive capacity at the expense of extensive and, therefore, ineffective rule sets. In this study, the proposed CORTEX method aims to produce smaller sets of rules with shorter rules without substantially decreasing its predictive performance. In the following section, the CORTEX method is briefly described.

\section{Cost-Sensitive Rule and Tree Extraction Method (CORTEX)}
Cost-sensitive learning is supervised learning that can alleviate the class imbalance problem in many applications. The class imbalance problem has been a recognized problem in the ML community for decades \citep{imbl_review}, and it can be defined as a problem where the number of samples across classes is not even. A popular measure of class imbalance is the class imbalance ratio, which in the binary classification framework can be expressed as the ratio of the majority class (the class with a more significant number of samples) to the minority class (the class with a smaller number of samples) size. In general, it is impossible to explicitly define an imbalance ratio that would deteriorate a classifier's performance due to other factors, such as small sample size or presence of sub-concepts, that could affect the performance \citep{imbl_review}.

Having a target variable with skewed class distribution can hinder the performance of many ML algorithms \citep{imbl_review}. The cost-sensitive learning method is one possible algorithm-level solution for tackling the class imbalance problem. The algorithm-level solutions are an alternative to data manipulation techniques such as oversampling or undersampling, with the idea of changing the internal structure of the learning algorithm to favor the minority class. One possible modification includes incorporating costs for incorrect classification (misclassification cost) for each class into the learning algorithm. 

The misclassification costs are typically represented as elements of a cost matrix. The cost matrix can be either class-dependent or sample-dependent, where the former means the misclassification costs are associated with the class. In contrast, the latter implies that each sample has its cost matrix defined. The assumption that the misclassification costs are constant across classes is more substantial and widespread through the application of most cost-sensitive learning algorithms \citep{Wang2021, Feng2015, Krawczyk2014, Lomax_2011, Sun_2006, Qin2005, Turney1995} since, in many real-life problems, the values in the matrix are unknown and not given by experts.  

Cost-Sensitive Rule and Tree Extraction (CORTEX) method is grounded in the multi-class cost-sensitive decision tree (CSDT) method. The CSDT method proposed by \citet{Correa_CSDT} is an ML algorithm for generating a tree model by considering a sample-dependent cost matrix during the tree-building procedure. The CSDT method belongs to the group of cost-sensitive learning methods \citep{Foundations_CSL}, that can be used in the imbalanced learning framework by considering class-dependent cost matrix i.e. matrix associated with classes instead of individual samples. Initially, the CSDT algorithm is proposed by \citet{Correa_CSDT} in the sample-dependent classification framework for imbalanced two-class classification problems in the financial domain. Here, the framework is extended to classification problems with more than two classes by inducing the concept of an n-dimensional class-dependent cost matrix into the CSDT algorithm. The CSDT algorithm is modified for a class-dependent framework and extended further to multi-class classification problems. 

In traditional decision tree learning algorithms, all samples are assumed to have equal importance. These algorithms assume a balanced class distribution and implicitly assume equal misclassification costs. The cost of misclassifying a sample is a function $\textbf{C}$ of the actual and predicted class, represented as a cost matrix:
\begin{equation}
    \textbf{C} = [C_{ij}]  = \begin{bmatrix}
                            C_{11} & C_{12} &  \dots  & C_{1K} \\
                            \vdots & \vdots & \ddots & \vdots & \\
                             C_{K1} & C_{K2} & \dots  & C_{KK}
                        \end{bmatrix}, \quad i,j = 1,\dots,K
\end{equation}
where $K$ represents number of classes, while $i$ and $j$ represent actual and predicted class, respectively. Accordingly, $C_{ij} = C(i,j)$ is the cost of predicting class $i$ when the actual (true) class is $j$.

The binary cost matrix is properly defined if two so-called "reasonable" conditions are satisfied \citep{Foundations_CSL}. The conditions imply that the cost of mislabeling a sample should always be greater than labeling it correctly. Violating conditions implies that one column dominates the other, and optimal prediction is the class corresponding to the dominated column. Intuitively, these conditions can be easily translated into a multi-class framework.

For binary classification tasks, the costs of making the error for the rare class are normally higher than those for the majority class. In literature, the cost is typically zero for correctly predicted outcomes, although the costs of correct classification can be non-zero \citep{Turney}.

Given a data represented as a collection of samples $\{x_{i}, y_{i}\}_{i=1}^{n}$ where $x_i$ is vector of features with length \textit{p} and $y_i \in \{1,\dots, K\}$ is a label (class) associated with the $i$th sample, and given a cost matrix $[C_{ij}]_{1<=i,j<=K}$, the CORTEX algorithm learns a function $f: X \rightarrow Y$, where $X \subseteq \mathbb{R}^{p}$ and $Y \subseteq \mathbb{R}$.

The learning phase consists of stratifying feature space into regions in a recursive manner (top-down greedy search). In the tree analogy, the regions are the nodes of the tree. The topmost node with only outgoing edges is known as a root node. Nodes with one incoming edge and two outgoing edges are called internal nodes. A terminal node (leaf node) has only one incoming edge and is denoted with the class label.

Unlike traditional decision tree, the CSDT \citep{Correa_CSDT} classifies sample $x_i$ in the region $R_m$ to the least costly class \textit{k(m)}:
\begin{equation}
    k(m) = \underset{k}{\operatorname{argmin}} \quad cost(f_k(m))
\end{equation}
where $f_k(m)$ is a function that assigns class label $k$ to all samples that belong to the node $m$ and $cost(f_k(m))$ is misclassification cost for class $k \in \{0,1,\dots,K\}$ at the node $m$, calculated as:
\begin{equation}
    cost(f_k(m)) = cost_{m}(k) = \sum_{i=1}^{K} N_{mi} C(i,k)
\end{equation}
Where $N_{mi}$ denotes the number of samples from class $i \in \{1,\dots, K\}$ in node $m$ and $C(i,k) = C_{ik}$ is cost of misclassifying sample from class $i$ into class $k \in \{1,\dots, K\}$, meaning the cost of predicting class $k$ in the node $m$ is equal to sum of misclassification costs for all samples in the node $m$ wrongly classified. Therefore, the terminal nodes are labeled to minimize the misclassification cost.

This approach is known as hard-labeling, meaning that a label assigned to each node is class, denoting the class membership of all samples in a given node. On the other hand, in the soft-labeling approach, a label assigned to each node is a class probability distribution. That is, each terminal node is labeled by a probability vector with a size corresponding to the number of classes. The class label is then determined based on the highest probability score. In our previous study \citep{kopanja}, soft-labeling is introduced into the original CSDT method to have probability membership for each sample and thereby have information about confidence in the prediction that the given sample belongs to a specific class. Labeling is introduced for binary frameworks by persevering the cost-dependence of labels. The formulation given in \citep{kopanja} is generalized for multi-class classification problems, leading to the proposed CORTEX method working for an arbitrary number of classes $K$. The mathematical formulation of cost-sensitive probabilities in CORTEX is given as:

\begin{equation}
    p_{mk} = \frac{1 - avgcost_{m}(f(k))}{\sum_{i = 1}^{K} avg cost(f(i))}
\end{equation}
where,
\begin{equation}
    avgcost_{m}(f(i)) = \frac{cost(f(i))}{\sum_{i = 1}^{K} avg cost(f(i))}
\end{equation}

Using the cost-sensitive probabilities defined above, classifying a sample in the least costly class is equivalent to classifying a sample in the class with the highest cost-sensitive probability. Notice that even if there were fewer samples from some class in the node, it could still be labeled as that class if the costs in the cost matrix are adequately defined. Therefore, the CORTEX model has a built-in bias towards the minority class(es), directed by misclassification costs defined in a cost matrix. 

The cost matrix can be manually defined and even tuned in settings where the number of classes is small. For example, in a binary setting, where costs for correct classification are considered to be equal to zero, and the misclassification cost for the majority class is set to be 1, the misclassification cost for the minority class can be observed as a hyperparameter of the method. However, with an increasing number of classes, it becomes computationally intractable to tune all misclassification costs in a matrix. On the other side, manually defining the values of a cost matrix could be tricky, considering the "reasonable" conditions the matrix needs to satisfy. Therefore, in the CORTEX method, the proposed default version of the cost matrix is defined by using class imbalance ratios among classes. Given the $N_i$ being the number of samples in class $i$, the values of a cost matrix are defined as: 
\begin{equation}
    C_{ij} = \frac{N_i + N_j}{N_i}
\end{equation}
where $C_{ij} = C(i,j)$ is the cost of wrongly classifying sample from class $i$ to class $j$ ($i,j = 1,\dots,K$) which reflects class imbalance ratio among the classes $i$ and $j$.

The default cost matrix is initially designed based on class imbalance ratios, providing a foundational approach to address the skewness of class distribution. However, future work could explore modifications of the cost matrix to enhance the proposed CORTEX's performance and make it more adaptable in a given scenario. A well-defined cost matrix can be considered an essential part of a cost-sensitive tree-building algorithm that should align with the specific objectives of the problem domain. Conversely, a poorly chosen cost matrix could lead to suboptimal results, such as bias toward majority classes or failure to imitate the behavior of the underlying model in the surrogate modeling. Careful consideration is essential when defining the cost matrix to ensure the algorithm achieves an intended balance between accuracy, fidelity, and other relevant measures depending on the problem domain.

\section{Experiment setup}
In the experimental setup of our study, the proposed CORTEX is used as a post-hoc XAI method by creating a surrogate tree model and automatically extracting a set of \textit{IF-THAN} rules from the obtained tree. The experiments were conducted on a set of eight publicly available datasets from the UCI Machine Learning Repository \footnote{https://archive.ics.uci.edu/}. Detailed dataset descriptions are given in Table \ref{data} below.

\begin{table}[h]
\centering
\scriptsize % Reduce font size further for compactness
\setlength{\tabcolsep}{3pt} % Reduce column padding
\renewcommand{\arraystretch}{1.1} % Adjust row height
\begin{tabular}{lccc}
\hline
\textbf{Dataset} & \textbf{Total number of samples} & \textbf{Total number of   features} & \textbf{Number of classes} \\
\hline
Abalone        & 4177  & 8   & 29 \\
Contraceptive  & 1473  & 9   & 3  \\
Credit         & 1000  & 20  & 2  \\
Mushroom       & 8124  & 22  & 2  \\
Page\_Block    & 5473  & 10  & 5  \\
Wave\_Form     & 5000  & 20  & 3  \\
Wine           & 6497  & 12  & 7  \\
Yeast          & 1484  & 8   & 10 \\
\hline
\end{tabular}
\caption{Description of the datasets.}
\label{data}
\end{table}

The datasets are briefly described in the study by~\citet{giulia_2020}, where selection is based on several criteria, including avoidance of curse of dimensionality, handcrafted features (both numerical and categorical), and categorical target variables with the number of classes ranging from two up to the 29 for the \textit{abalone} dataset.

\begin{figure*}[h!]
    \centering
    \includegraphics[width=1.0\linewidth]{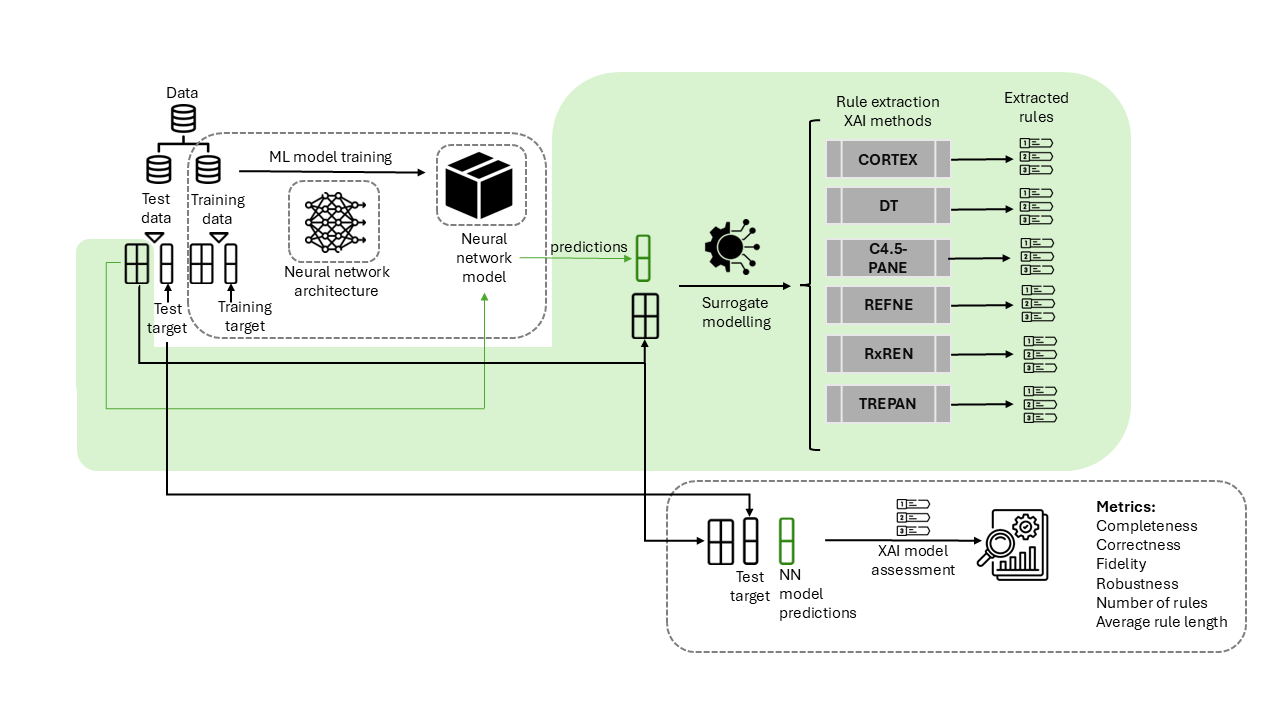}
    \caption{Experimental study design for creating and evaluating XAI surrogate models using six rule extractors.}
    \label{diagram}
\end{figure*}

The experiment was designed as shown in the diagram below (Figure \ref{diagram}). The first step of the experimental setup is the data preparation process, where datasets with nominal features are encoded using a one-hot encoding technique. Then, the neural network models are trained on 70\% of data using two different architectures in terms of the number of layers, the first being the vanilla feed-forward neural network with a single hidden layer and the other with two fully connected hidden layers. Afterward, the post-hoc surrogate models are created using test data and predictions given by the neural network. The cost-sensitive tree model is created using the CORTEX algorithm, and a set of if-then rules is automatically generated from the obtained tree model. 

Six metrics were selected to assess the degree of explainability of the rule sets, including completeness, correctness, fidelity, robustness, number of rules, and average rule length. Completeness is calculated as the ratio of samples covered by rules over the total number of samples. Correctness refers to the proportion of samples correctly classified by the rules relative to the total number of samples. Fidelity, on the other hand, is the proportion of samples for which the predictions made by both the neural networks and the rules align compared to the total number of samples. Robustness is the proportion of the difference between the predictions made by a set of rules on test samples with added Gaussian noise and the predictions made on the original test samples relative to the total number of test samples. 
The cardinality of the rule set is expressed as a number of rules and the average number of antecedents connected with the AND operator in a rule across all rules (expressed as the average rule length). 
Formal definitions of these measures can be found in studies \citet{giulia_2020} and \cite{giulia_2021}.

Five rule exaction methods are used to provide a comprehensive comparison analysis. The abundance of XAI rule-extraction methods makes it computationally extensive to use them all in comparison analysis. Four rule-extractors, C4.5-PANE, REFNE, RxREN, and TREPAN, are extensively studied in the literature~\citep{giulia_2020, giulia_2021} and therefore considered as a strong baseline for evaluating the CORTEX method proposed in this study. Furthermore, considering the proposed rule method is a tree-based algorithm, the selected subset of benchmarking methods is extended with a traditional decision tree classifier (DT) to provide a more comprehensive evaluation. Moreover, due to the imbalanced distribution of the target variable, the DT method is trained by considering the class imbalance by automatically adjusting weights to be inversely proportional to class frequencies in the weighted impurity gain measure used in the feature splitting criterion \citep{DTw}. Without considering the imbalance in DT model building, the rule sets extracted from it would not be reasonably comparable with rules from the CORTEX since the CORTEX naturally considers the class imbalance ratio in data. The weighted decision tree method (DT) is implemented using scikit-learn \citep{scikit-learn}. The algorithms for other rule-extractors are obtained from \url{https://github.com/giuliavilone/rule_extractor}.

The final step of the experimental process involves ranking the selected XAI methods based on the six evaluation metrics. For this purpose, we used the Friedman test \citep{Friedman1, Friedman2} and Wilcoxon signed-rank test \citep{Wilcoxon}, non-parametric tests that discover whether rule-extractors perform equally. The process is conducted using 100-time repeated results for reliability and reproducibility for both setups (both types of architectures of neural networks). 

\section{Results and discussion}

This section presents results from two different experimental setups, differing in the neural network's architecture. The first experiment is conducted on the vanilla feed-forward neural networks with a single hidden layer (NN-1), and the second experiment is carried out on the feed-forward neural network with two fully connected hidden layers (NN-2). For all hyperparameters of the networks, such as the number of hidden neurons, activation function, dropout rate, optimization algorithm, and others, optimal values are obtained from Table 2 reported by~\cite {giulia_2020}. The training process was early-stopped to prevent overfitting. Accuracy for both architectures of neural network models on training and test sets is given in Table~\ref{nn_results}. The results in the table display the average accuracy score across 100 runs. Notably, depending on the dataset, a more complex neural network (NN-2) does not necessarily improve the accuracy score. However, for some datasets, including \textit{abalone}, \textit{credit}, \textit{wave-form}, and \textit{yeast}, the accuracy score on the test set is slightly increased.

\begin{table*}[h!]
\centering
\caption{Performance evaluation of neural network models with different architectures across different datasets.}
\begin{tabular}{clllllllll}
\hline
\multirow{12}{*}{*} \\
\multicolumn{1}{l}{Model} & \multicolumn{1}{c}{Metric} & \multicolumn{8}{c}{Dataset}                                                                                                                    \\ \cline{3-10}
\multicolumn{1}{c}{} & \multicolumn{1}{c}{} & \multicolumn{1}{l|}{abalone} & \multicolumn{1}{l|}{contraceptive} & \multicolumn{1}{l|}{credit} & \multicolumn{1}{l|}{mushroom} & \multicolumn{1}{l|}{page\_blocks} & \multicolumn{1}{l|}{wave\_form} & \multicolumn{1}{l|}{wine} & \multicolumn{1}{l}{yeast} \\ \hline
\multirow{2}{*}{NN-1} & Acc\_train & 0.272	& 0.576 &	0.746 &	1.000	& 0.966 &	0.769 &	0.516 &	0.595 \\ & Acc\_test  & 0.263 &	0.549 &	0.712 &	1.000 &	0.963 &	0.769 &	0.511 &	0.580 \\
\multirow{2}{*}{NN-2}   & Acc\_train & 0.277   & 0.566    & 0.769   & 1.000    & 0.965                             & 0.791 & 0.493 & 0.602 \\ & Acc\_test  & 0.267  & 0.540  & 0.739  & 1.000  & 0.962 & 0.789 & 0.486 & 0.584                     
\end{tabular}
\begin{tablenotes}
    \item[]\textit{Note: NN-{$k$} is feed-forward neural network model trained with $k$ hidden layers.}
    \end{tablenotes}
  \label{nn_results}%
\end{table*}

In the next step of experimental design, six XAI rule-based methods are trained given a test dataset and trained neural network model. Afterward, a set of rules is extracted from each XAI model. Then, the sets of produced rules are evaluated using six quantitative measures of the degree of explainability. The objective evaluation is accomplished by excluding any human intervention. By combining the completeness, correctness, faithfulness (fidelity), and robustness of a rule set, we can effectively estimate the validity of the representation of an underlying model’s inferential process. 
The rule set should appropriately classify any sample, be faithful to the underlying model, and produce inferences that will not vary when inputs are slightly distorted by applying Gaussian noise. Two other metrics, the number of rules and average rule length, are used to assess the syntactic simplicity of the rules. Both of these measures should be minimized for a given rule set to ensure the rules are easily interpretable and understandable by humans.

\begin{figure*}[ht!]
    \centering
    \includegraphics[width=0.9\linewidth]{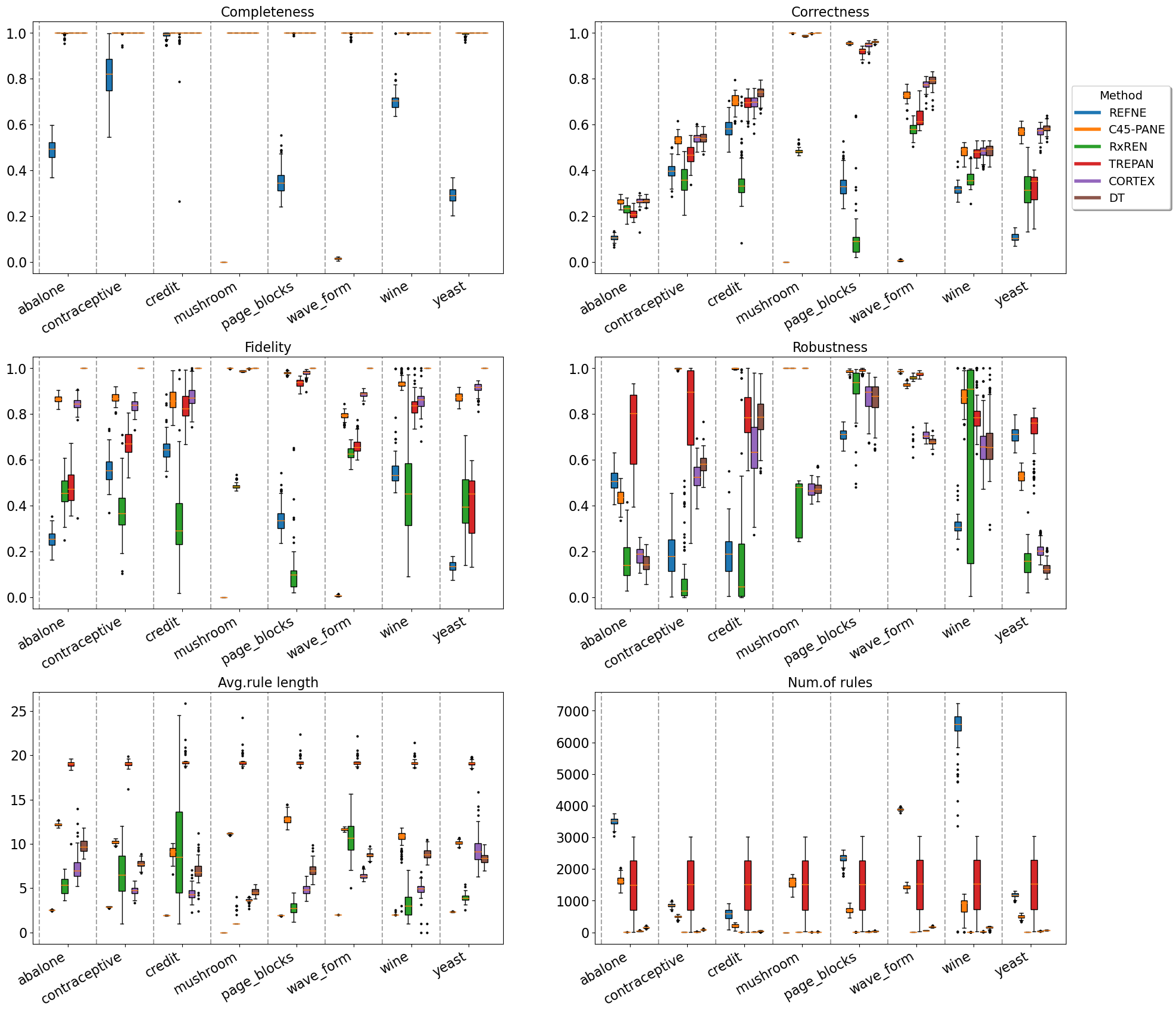}
    \caption{Distribution of six quantitative metrics for evaluating rule-extractor XAI methods across several datasets.}
    \label{all_metrics_datasets_main_text}
\end{figure*}

Results for the six quantitative measures for all rule-extractors generated for NN-2 models across datasets are shown in Figure~\ref{all_metrics_datasets_main_text}. Each box plot visually summarizes the distribution of a metric across 100 runs for a given method and dataset. A more granular representation of the results for each dataset separately is given in Appendix 1. 

The analysis of the reported results leads to several conclusions. The rule sets obtained from all methods except REFNE and RxREN cover all input samples. Therefore, the proposed CORTEX produces rule sets that reach 100\% of completeness throughout all the datasets, ensuring that the rules cover every sample in a given dataset. The same conclusion about REFNE failing to provide a complete set of rules is also observed in other studies \citep {giulia_2020, giulia_2021}. REFNE's underperformance is due to its limitation of maximally three features in a rule, which may not be sufficient to cover the entire feature space for complex datasets.

\begin{table*}[h!]
\scriptsize
\centering
\begin{tabular}{p{1.5cm}p{1.5cm}llllll}
\multicolumn{1}{c}{\multirow{2}{*}{Metric}} & \multicolumn{1}{c}{\multirow{2}{*}{Dataset}} & \multicolumn{6}{c}{\normalsize {XAI rule-extraction   methods}} \\
\multicolumn{1}{c}{} & \multicolumn{1}{c}{} & REFNE & C4.5-PANE & RxREN & TREPAN & CORTEX & DT \\ \hline
&  & MEAN \( \left( \text{STD} \right) \) & MEAN (STD) & MEAN (STD) & MEAN (STD) & MEAN (STD) & MEAN (STD) \\ 
\multirow{8}{*}{Completeness} & abalone & 0.488 (0.046) & 1.0 (0.0) & 0.998 (0.007) & 1.0 (0.0) & \cellcolor{Gray!20} 1.0 (0.0) & 1.0 (0.0) \\
 & contraceptive & 0.82 (0.094) & 1.0 (0.0) & 0.999 (0.008) & 1.0 (0.0) & \cellcolor{Gray!20} 1.0 (0.0) & 1.0 (0.0)  \\
 & credit & 0.992 (0.013) & 1.0 (0.0) & 0.988 (0.077) & 1.0 (0.0) & \cellcolor{Gray!20} 1.0 (0.0) & 1.0 (0.0) \\
 & mushroom & 0.0 (0.0) & 1.0 (0.0) & 1.0 (0.0) & 1.0 (0.0) & \cellcolor{Gray!20} 1.0 (0.0) & 1.0 (0.0) \\
 & page\_blocks & 0.353 (0.062) & 1.0 (0.0) & 1.0 (0.002) & 1.0 (0.0) & \cellcolor{Gray!20} 1.0 (0.0) & 1.0 (0.0) \\
 & wave\_form & 0.014 (0.004) & 1.0 (0.0) & 0.998 (0.007) & 1.0 (0.0) & \cellcolor{Gray!20} 1.0 (0.0) & 1.0 (0.0) \\
 & wine & 0.706 (0.054) & 1.0 (0.0) & 1.0 (0.0) & 1.0 (0.0) & \cellcolor{Gray!20} 1.0 (0.0) & 1.0 (0.0) \\
 & yeast & 0.287 (0.036) & 1.0 (0.0) & 0.996 (0.007) & 1.0 (0.0) & \cellcolor{Gray!20} 1.0 (0.0) & 1.0 (0.0) \\ \hline
\multirow{8}{*}{Correctness} & abalone & 0.105 (0.012) & 0.265 (0.013) & 0.23 (0.023) & 0.209 (0.018) & \cellcolor{Gray!20} 0.265 (0.018) & 0.267 (0.013 \\
 & contraceptive & 0.397 (0.039) & 0.532 (0.023) & 0.358 (0.063) & 0.465 (0.042) & \cellcolor{Gray!20} 0.541 (0.025) & 0.54 (0.024) \\
 & credit & 0.585 (0.043) & 0.702 (0.032) & 0.353 (0.093) & 0.693 (0.036) & \cellcolor{Gray!20} 0.695 (0.033) & 0.739 (0.028) \\
 & mushroom & 0.0 (0.0) & 1.0 (0.0) & 0.485 (0.012) & 0.988 (0.002) & \cellcolor{Gray!20} 0.999 (0.001) & 1.0 (0.0) \\
 & page\_blocks & 0.337 (0.058) & 0.955 (0.005) & 0.106 (0.1) & 0.92 (0.015) & \cellcolor{Gray!20} 0.948 (0.014) & 0.962 (0.005) \\
 & wave\_form & 0.006 (0.002) & 0.727 (0.023) & 0.578 (0.027) & 0.631 (0.044) & \cellcolor{Gray!20} 0.773 (0.021) & 0.789 (0.027) \\
 & wine & 0.318 (0.025) & 0.482 (0.026) & 0.367 (0.044) & 0.474 (0.027) & \cellcolor{Gray!20} 0.479 (0.029) & 0.486 (0.027) \\
 & yeast & 0.107 (0.018) & 0.57 (0.022) & 0.314 (0.079) & 0.328 (0.053) & \cellcolor{Gray!20} 0.568 (0.025) & 0.584 (0.019) \\ \hline
\multirow{8}{*}{Fidelity} & abalone & 0.252 (0.036) & 0.866 (0.017) & 0.458 (0.071) & 0.481 (0.073) & \cellcolor{Gray!20} 0.841 (0.056) & 1.0 (0.0)  \\
 & contraceptive & 0.555 (0.063) & 0.871 (0.022) & 0.375 (0.1) & 0.669 (0.062) & \cellcolor{Gray!20} 0.836 (0.028) & 1.0 (0.0) \\
 & credit & 0.651 (0.059) & 0.862 (0.05) & 0.331 (0.169) & 0.832 (0.067) & \cellcolor{Gray!20} 0.875 (0.051) & 1.0 (0.0) \\
 & mushroom & 0.0 (0.0) & 1.0 (0.0) & 0.485 (0.012) & 0.988 (0.002) & \cellcolor{Gray!20} 0.999 (0.001) & 1.0 (0.0) \\
 & page\_blocks & 0.343 (0.06) & 0.98 (0.005) & 0.111 (0.103) & 0.936 (0.017) & \cellcolor{Gray!20} 0.979 (0.014) & 1.0 (0.0) \\
 & wave\_form & 0.006 (0.002) & 0.794 (0.018) & 0.628 (0.029) & 0.667 (0.038) & \cellcolor{Gray!20} 0.888 (0.012) & 1.0 (0.0) \\
 & wine & 0.554 (0.085) & 0.937 (0.021) & 0.469 (0.2) & 0.834 (0.049) & \cellcolor{Gray!20} 0.857 (0.044) & 1.0 (0.0) \\
 & yeast & 0.134 (0.022) & 0.872 (0.021) & 0.411 (0.122) & 0.409 (0.12) & \cellcolor{Gray!20} 0.913 (0.025) & 1.0 (0.0) \\ \hline
\multirow{8}{*}{Robustness} & abalone & 0.512 (0.046) & 0.435 (0.039) & 0.164 (0.085) & 0.739 (0.169) & \cellcolor{Gray!20} 0.18 (0.038) & 0.148 (0.039) \\
 & contraceptive & 0.18 (0.094) & 0.998 (0.002) & 0.081 (0.127) & 0.793 (0.223) & \cellcolor{Gray!20} 0.528 (0.06) & 0.581 (0.047) \\
 & credit & 0.192 (0.098) & 0.999 (0.002) & 0.167 (0.253) & 0.8 (0.1) & \cellcolor{Gray!20} 0.649 (0.146) & 0.785 (0.096)\\
 & mushroom & 1.0 (0.0) & 1.0 (0.0) & 0.392 (0.117) & 1.0 (0.0) & \cellcolor{Gray!20} 0.472 (0.031) & 0.476 (0.032) \\
 & page\_blocks & 0.709 (0.029) & 0.985 (0.007) & 0.916 (0.094) & 0.989 (0.008) & \cellcolor{Gray!20} 0.873 (0.069) & 0.863 (0.078) \\
 & wave\_form & 0.986 (0.004) & 0.929 (0.008) & 0.943 (0.067) & 0.975 (0.008) & \cellcolor{Gray!20} 0.707 (0.017) & 0.68 (0.016) \\
 & wine & 0.327 (0.105) & 0.878 (0.056) & 0.644 (0.407) & 0.803 (0.089) & \cellcolor{Gray!20} 0.665 (0.102) & 0.676 (0.12) \\
 & yeast & 0.713 (0.036) & 0.528 (0.028) & 0.152 (0.064) & 0.725 (0.091) & \cellcolor{Gray!20} 0.205 (0.033) & 0.126 (0.031) \\ \hline
\multirow{8}{*}{Avg.rule   length} & abalone & 2.511 (0.035) & 12.177 (0.184) & 5.272 (0.938) & 18.912 (0.937) & \cellcolor{Gray!20} 7.268 (1.41) & 9.781 (0.784) \\ & contraceptive & 2.868 (0.047) & 10.185 (0.184) & 6.63 (2.568) & 19.014 (0.375) & \cellcolor{Gray!20} 4.7 (0.498) & 7.755 (0.391) \\
 & credit & 1.925 (0.025) & 9.004 (0.656) & 9.45 (5.855) & 19.296 (0.779) & \cellcolor{Gray!20} 4.361 (0.701) & 6.906 (1.152) \\
 & mushroom & 0.0 (0.0) & 11.165 (0.069) & 1.145 (0.519) & 19.248 (0.618) & \cellcolor{Gray!20} 3.642 (0.194) & 4.581 (0.385) \\
 & page\_blocks & 1.934 (0.027) & 12.793 (0.55) & 2.79 (0.73) & 19.181 (0.432) & \cellcolor{Gray!20} 4.853 (0.592) & 7.009 (0.772) \\
 & wave\_form & 2.004 (0.001) & 11.629 (0.127) & 10.787 (2.019) & 19.188 (0.419) & \cellcolor{Gray!20} 6.42 (0.367) & 8.754 (0.294) \\
 & wine & 2.011 (0.064) & 10.55 (1.414) & 3.027 (1.463) & 19.149 (0.353) & \cellcolor{Gray!20} 4.857 (0.856) & 8.713 (1.313) \\
 & yeast & 2.319 (0.029) & 10.102 (0.237) & 3.897 (0.409) & 19.084 (0.251) & \cellcolor{Gray!20} 9.409 (1.656) & 8.342 (0.511)\\ \hline
\multirow{8}{*}{Num.of   rules} & abalone & 3507.91 (138.819) & 1618.16 (153.927) & 7.49 (0.772) & 1505.25 (885.065) & \cellcolor{Gray!20} 41.71 (11.297) & 156.95 (17.868)\\
 & contraceptive & 851.13 (53.993) & 491.56 (45.441) & 3.0 (0.0) & 1511.51 (885.088) & \cellcolor{Gray!20} 18.04 (4.257) & 85.94 (9.77) \\
 & credit & 569.52 (181.353) & 204.95 (56.758) & 1.98 (0.141) & 1515.18 (885.14) & \cellcolor{Gray!20} 12.23 (2.785) & 37.99 (10.785) \\
 & mushroom & 0.0 (0.0) & 1574.94 (175.449) & 2.0 (0.0) & 1517.18 (885.14) & \cellcolor{Gray!20} 10.13 (0.906) & 16.42 (2.606)\\
 & page\_blocks & 2326.3 (164.93) & 693.63 (96.224) & 4.18 (0.557) & 1520.24 (885.162) & \cellcolor{Gray!20} 17.83 (3.851) & 34.26 (7.168) \\
 & wave\_form & 3892.77 (31.813) & 1433.69 (64.53) & 3.0 (0.0) & 1524.08 (885.112) & \cellcolor{Gray!20} 52.76 (5.772) & 174.61 (12.44) \\
 & wine & 6307.88 (1143.619) & 805.04 (300.82) & 1.92 (0.367) & 1526.86 (885.217) & \cellcolor{Gray!20} 21.83 (6.76) & 143.65 (40.835) \\
 & yeast & 1179.57 (73.511) & 494.24 (56.023) & 8.41 (0.653) & 1529.86 (884.985) & \cellcolor{Gray!20} 37.33 (6.014) & 65.58 (6.86)
\end{tabular}
\caption{Summary of results of six evaluation measures across eight datasets for six rule-based methods. The table shows the means and standard deviations.}
\label{table_results}
\end{table*}

All tree-based models, including the proposed CORTEX, stand out as top-performing methods regarding correctness and fidelity across all datasets. However, regarding robustness, CORTEX is underperforming compared to other tree-based methods, while it is competitive with other rule extractors. Therefore, CORTEX can be enhanced to ensure that minor input modifications do not significantly affect the model's predictions. The CORTEX method demonstrates advantages over other methods (excluding RxREN) based on the number of rules generated and the average length of those rules. It can produce smaller rule sets with shorter average rules across datasets that vary in both the number of classes and class imbalance ratios. Notably, CORTEX outperforms the top-ranked C4.5-PANE method from the study by~\cite {giulia_2020} in terms of both the number of rules and the average rule length. Furthermore, the plots indicate that CORTEX is more stable than the TREPAN method regarding the number of rules. 

The resulting explanations of CORTEX are easy-to-understand \textbf{IF-THEN} rules, as illustrated in Figure ~\ref{rules} given in the Appendix for the credit dataset. The conditions (antecedents) of each rule are enclosed in parentheses $"( )"$ and connected using the $"AND"$ operator, with the class label specified after the $"THEN"$ keyword. As shown in Table~\ref{table_results} the average number of rules produced by CORTEX for the credit dataset is 12, and the average number of antecedents in a rule set is 4. For the credit dataset, CORTEX ranks second in the number of rules, following RxREN. When it comes to average rule length, CORTEX is behind REFNE. Despite this, CORTEX outperforms both REFNE and RxREN in all other metrics. While CORTEX may not have the lowest average rule length nor the smallest set of rules, it demonstrates superior performance considering various aspects compared to its counterparts. Therefore, CORTEX achieves a strong balance between different metrics, demonstrating effective performance across various desirable properties.

To assess whether a certain rule-extraction method performs significantly differently, we employ a non-parametric test. This choice is made because non-parametric tests do not rely on assumptions such as normal distribution or homogeneity, which are prerequisites for parametric tests. We selected the Friedman test \citep{Friedman1, Friedman2} as a non-parametric statistical method to identify statistically significant differences among six rule-based methods across several runs for each of the six metrics and datasets. The results of the test (whether we accept or reject the null hypothesis, which states there is no significant difference) are evaluated using a significance level of 0.05. We seek to verify if there are statistically significant differences in the degree of explainability (measured with six different metrics) of rule sets extracted by the six rule extractors.

The results for both setups, NN-1 and NN-2 models, are identical and indicate that there are statistically significant differences among the rule-extraction methods across all metrics, except for completeness. This applies to all datasets except for the \textit{mushroom}  dataset, where we do not reject the hypothesis for three measures of agreement between the rule explainer and the machine learning model: correctness, fidelity, and robustness. A more detailed examination of the evaluation metrics for the \textit{mushroom}  dataset can be found in Appendix. For the number of rules and the average rule length, the tests show statistically significant differences in the degree of explainability of the rule sets extracted by six methods across all datasets.

In the next step, we conduct a nonparametric test to compare pairs of rule-extractors. The \citep{Wilcoxon} is employed to assess whether the medians of the results from two methods differ, helping us determine if compared methods perform equally. Based on the test results and chosen significance level of 0.05, it can be concluded that there is no significant difference in the performance of the two models being compared if the null hypothesis is accepted. We will briefly describe the results obtained for the second setup using the neural network model with two hidden layers (NN-2). Similar results were found with another setup using vanilla neural networks with one hidden layer (NN-1).

In terms of the completeness metric, the results show statistically significant differences in the rule sets generated by CORTEX when compared to both REFNE and RxREN across all datasets. The null hypothesis can be rejected for all pairs of methods and datasets regarding the two simplicity measures of the rule sets: the number of rules and the average rule length. There is also a notable difference in the robustness metrics between rules extracted from CORTEX and those derived from other methods, with the exception of the \textit{page-blocks} dataset and the C4.5-PANE method. Regarding robustness, no significant difference is observed in the rule sets derived from CORTEX and DT for the \textit{mushroom}, \textit{page blocks}, and \textit{wine} datasets. Additionally, for the \textit{wine} dataset, the test does not reject the null hypothesis for the pair of CORTEX and RxREN in terms of robustness. When examining the correctness and fidelity metrics, it is observed that the hypothesis is not rejected in several cases, though the results vary depending on the dataset and the method compared to CORTEX (for the NN-1 model, there are fewer exceptions than for the NN-2 model). Exceptions for the correctness metric include CORTEX and C4.5-PANE on the \textit{abalone} dataset, CORTEX and DT on the \textit{contraceptive} dataset, CORTEX and TREPAN on the \textit{credit} dataset, and CORTEX and C4.5-PANE on the \textit{yeast} dataset. 

The final step of the experimental process involves ranking the selected XAI methods based on the six evaluation metrics. The process is conducted using 100-time repeated results with randomized splits of the training and test sets to ensure reliability and reproducibility. First, we thoroughly investigated the obtained rankings for each run by summing all ranks per metric across eight datasets. To provide a more comprehensive analysis in a concise format, the following comparative table (Table \ref{comp_table}), which goes beyond individual analysis per dataset, in a certain way summarizes the results given in Table \ref{table_results}. In Table \ref{comp_table}, the best method is determined by identifying the rule extractor with the highest average rank, calculated as the mean value of ranks across all datasets for each metric. Similarly, the rule extractor is deemed worst if it has the lowest average rank across all datasets. The third column displays the average rank of CORTEX for given metrics across all datasets. The fourth column presents the results of the winning method in the comparative evaluation between CORTEX and DT for the specified metrics across all datasets. Additionally, color coding indicates whether there are statistically significant differences in the average performance of the two methods based on the results of the Wilcoxon statistical test.

\begin{table*}[h!]
    \centering
    \begin{tabular}{ccccc}
        ~ & \textbf{Best method} & \textbf{Worst method} & \textbf{CORTEX rank} & \textbf{CORTEX vs. DT} \\ \hline
        Completeness & C4.5-PANE, TREPAN, CORTEX, DT & REFNE & 1 & CORTEX, DT \\ \hline
        Correctness & DT & REFNE & 3 & \cellcolor{Gray!20} DT \\ \hline
        Fidelity & DT & REFNE & 3 & \cellcolor{Gray!20} DT \\ \hline
        Robustness & TREPAN & RxREN & 4 & \cellcolor{Gray!20} CORTEX \\ \hline
        Avg.rule length & REFNE & TREPAN & 3 & \cellcolor{Gray!20} CORTEX \\ \hline
        Num.of rules & RxREN & TREPAN & 2 & \cellcolor{Gray!20} CORTEX \\ \hline
    \end{tabular}
\caption{Results of comparative analysis of six evaluation measures for six rule-based methods.}
\begin{tablenotes}
    \item[]\textit{Note: Color coding indicates whether there are statistically significant differences in the average performance of the CORTEX and DT methods based on the results of the Wilcoxon signed-rank test.}
    \end{tablenotes}
\label{comp_table}
\end{table*}

The results presented in Table \ref{comp_table} indicate that CORTEX is not the top-performing method when considering the number of rules or the average rule length. In these categories, CORTEX ranks as either the second or third best, as seen in the third column labeled "CORTEX rank." However, it does perform better than the DT method in both of these measures of explainability, and the difference in performance between these two methods is statistically significant on average, according to the Wilcoxon test. Conversely, while REFNE and RxREN excel in the two measures of explainability, they rank poorly in the other four metrics. As shown in the second column, REFNE is ranked the lowest in terms of completeness, correctness, and fidelity, while RxREN is identified as the least robust method. Notably, CORTEX surpasses DT in terms of robustness, even though it is not the most robust method. While TREPAN is recognized as the most robust rule extractor, it ranks poorly concerning the number of rules and average rule length.

The table illustrates the trade-off between the simplicity of rules—measured by the number of rules and the average rule length—and the other four metrics. The same observation has been made in other studies \citep{giulia_2020, giulia_2021}. Notably, the CORTEX achieves a good balance among the desirable qualities of XAI methods, as it successfully reduces both the number of rules and the average number of antecedents per rule. However, maintaining these two measures at satisfactory levels—compared to underperforming baselines in the literature—comes at the cost of slightly decreased correctness and fidelity, though the decline is not drastic. Furthermore, CORTEX demonstrates better robustness than DT, which is considered the best method for correctness and fidelity. Importantly, completeness, which refers to the coverage of all samples with rules extracted from CORTEX, remains at 100\%. This is in contrast to the best-ranked baselines, REFNE and RxREN, which do not achieve full coverage despite performing well in terms of average rule length and number of rules.

\begin{figure}[h!]
    \centering
    \includegraphics[width=0.8\linewidth]{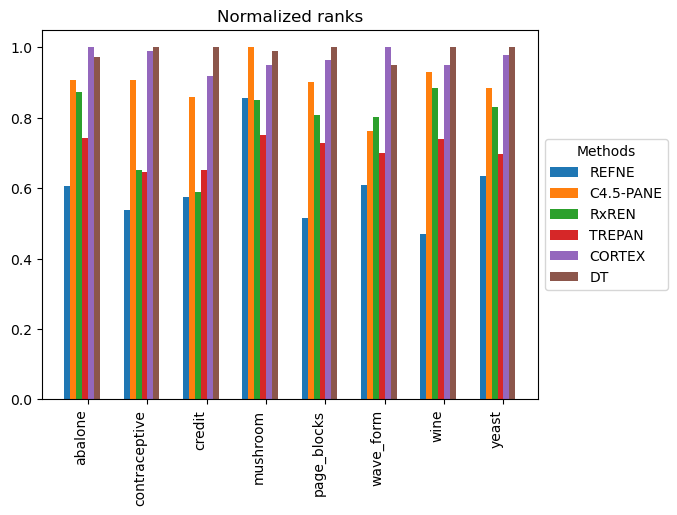}
    \caption{Normalized ranks for six rule extraction XAI methods across datasets.}
    \label{ranks}
\end{figure}

The final ranking of the methods is conducted for each run and then summed for each metric across all datasets. The total rank is normalized, and the results are displayed in Figure \ref{ranks} for the NN-2 model. Similar ranking is obtained for the NN-1 model setup. Normalization of the ranking is performed due to the aggregation of ranks across multiple metrics grouped by datasets.

As can be observed, the CORTEX and DT are rule extraction methods ranked consistently higher across all datasets, excluding the \textit{mushroom} dataset where the C4.5-PANE holds the highest rank. The results indicate that tree-based rule extraction methods (CORTEX, DT and C4.5-PANE)  for explainability are the top three choices for explainability across all datasets.This suggests that using a tree-based algorithm is preferable to using other rule-based extractors for rule generation. Additionally, the CORTEX is ranked higher than DT for the \textit{abalone} dataset, which is the dataset with the most number of classes and, therefore, the most challanging problem for ML model training (with the accuracy less than $30\%$, as confirmed by the performance evaluation of the neural network models reported in Table \ref{nn_results}). For the \textit{wave-form} dataset, the CORTEX also achieves the highest rank. Nonetheless, for other datasets (\textit{contraceptive}, \textit{wine}, and \textit{yeast}), CORTEX's ranking is competitive with rank of DT, as its rank does not significantly fall below that of the DT model.

Tree-based models consistently rank highest across various datasets, suggesting they excel at generating interpretable rule sets for complex predictive tasks. These models can be effectively trained on diverse datasets without significantly compromising other general evaluation metrics. Our findings also confirm the trade-off observed in previous studies (\cite{giulia_2020}, \cite{giulia_2021}) between the size of rule sets—measured by the number of rules and antecedents—and four key metrics: completeness, correctness, fidelity, and robustness. Notably, CORTEX outperforms the top-ranked method, C4.5-PANE, in \citet{giulia_2020}, particularly on highly unbalanced datasets. This demonstrates CORTEX’s ability to effectively identify cases where neural networks mistakenly classify minority samples as part of the majority class, while also producing more understandable rules without a significant decline in the other four metrics.

The proposed CORTEX method demonstrates competitive performance compared to other tree-based models, showcasing its effectiveness in handling black-box models on diverse datasets. Furthermore, it surpasses the capabilities of inherent rule-extraction techniques, delivering superior results in terms of correctness and fidelity. The extracted rule sets suggest the advantages of using the CORTEX method over alternative methods. Specifically, CORTEX can produce smaller rule sets with, on average, shorter rules across datasets with a diverse number of classes. However, this advantage comes with the trade-off of having a model that is less accurate and robust, although it effectively balances this trade-off. Overall, the results underscore the potential of CORTEX as a powerful XAI tool for scenarios that require the generation of clear, human-understandable rules while maintaining good predictive performance.

\section{Conclusion}
Recent advancements in technology and AI-driven systems have raised interest among various stakeholders and regulatory entities in understanding these systems. There is a growing demand for trustworthy AI, especially in practical applications where AI-driven models could endanger human lives. It is worth noting that existing XAI techniques do not fully address all the concerns surrounding explanations and there is a clear need for new XAI methods capable of handling complex ML models in sophisticated real-world applications. Promising approach for ensuring trustworthiness, reliability, transparency, and fairness in black-box models include post-hoc model-agnostic surrogate models. Extracting explanations in the form of trees or rules are strategies employed by certain global, post-hoc, model-agnostic XAI methods to enhance the explainability of an ML model. Decision trees and rule sets as two different representation types of explanations, may be favored in different contexts.
This study presents a novel CORTEX method, based on the cost-sensitive decision tree (CSDT) algorithm. The  CSDT is adapted for classification problems that involve more than two classes by incorporating an n-dimensional class-dependent cost matrix into the CSDT algorithm. The CSDT algorithm is modified for a class-dependent framework and further extended to handle multi-class classification problems.  The tree-based explanation model generated from CORTEX is coverted into a set of rules. This study provides a comprehensive comparision of CORTEX against other XAI rule-extraction methods on two architectures of neural network models as the underlying model. The feed-forward neural networks are configured with one or two hidden layers on eight publicly available datasets within the classification framework. To determine the best surrogate model, quantitative evaluation is conducted by identifying six objective metrics to assess the degree of explainability of the rule sets extracted from the six methods being compared. Namely, the metrics include completeness, correctness, fidelity, robustness, number of rules, and average rule length. These metrics assess the degree of agreement of surrogate rule-based models with the underlying model as well as the syntactic simplicity of rule sets generated by the surrogate methods. Our findings confirm a trade-off between the two metrics evaluating the syntactic simplicity of the rule sets (number of rules and average rule length) generated by the CORTEX and the other four metrics. This trade-off is also noted in other studies for other rule extractors used as surrogate models.
The Friedman test is employed to determine if there are statistically significant differences in the performance of the six XAI methods under consideration. The results indicate that the null hypothesis can be rejected for all metrics except for completeness. For this metric, none of the methods consistently outperformed the others across all datasets. However, the correctness, fidelity, and robustness metrics showed that some methods outperformed others for specific dataset.
To compare pairs of rule-extractors, the Wilcoxon signed-rank test is conducted. For both measures of the simplicity of the rule sets—the number of rules and the average rule length—the results reveal statistically significant differences between the rule sets generated by CORTEX and the other rule-extractors across all datasets. In terms of completeness, the null hypothesis is rejected for some pairs, but it is not rejected for the CORTEX and REFNE across all datasets. The results for other tree metrics indicate that CORTEX performance is statistically significantly different from the other methods, with a few exceptions. 
The final evaluation step involved ranking the selected methods based on six evaluation metrics. First, we ranked the methods for each metric across multiple runs of the selected rule-based surrogate methods, resulting in six different rankings. These ranks were then normalized and summed across datasets to obtain a single final ranking for each method. According to the results, tree-based rule extraction methods for explainability are consistently ranked higher across the datasets. Specifically, CORTEX is ranked highest for the dataset with the highest number of classes and the most complex problem for ML model training. For other datasets, CORTEX is ranked as the best method or second-best, while the balanced DT method is ranked the highest in some cases. Our findings demonstrate that CORTEX, as a novel rule-based, post-hoc, model-agnostic XAI method, outperforms other baseline rule extractors found in the literature. The CORTEX also has competitive performance with the balanced DT method, which was included as an additional benchmark rule-extractor method due to its similar characteristics (the tree structure of the model) with CORTEX method. More importantly, the preference towards CORTEX can be justified by the simplicity of rule sets, which scale more efficiently and are more sustainable compared to more complex alternatives. While CORTEX may not have the lowest average rule length nor the smallest set of rules, it demonstrates superior performance considering various aspects compared to its counterparts. Therefore, CORTEX achieves a strong balance between different metrics, demonstrating effective performance across various desirable properties.
Future work will expand the experimental evaluation to include additional datasets that vary in the number of classes and class imbalance ratios. Various neural network architectures will be examined based on the complexity of new datasets. Additionally, more rule-extraction methods could be incorporated for a more comprehensive comparison. The assessment of the rules extracted from different methods could also take into account the consistency of explanations, examining similarity of various rule sets generated for the same problem. 
To enhance the reliability of CORTEX, the algorithm could be refined so that minor input modifications do not lead to significant changes in the model's predictions. This robustness improvement could foster greater confidence in the model's stability and efficacy, making it a more trustworthy tool for practical applications.

\bibliographystyle{apalike}  
\bibliography{references}  %%% Remove comment to use the external .bib file (using bibtex).

\begin{thebibliography}{}

\bibitem[Ali et~al., 2023]{ali}
Ali, S., Abuhmed, T., El-Sappagh, S., Muhammad, K., Alonso-Moral, J.~M., Confalonieri, R., Guidotti, R., {Del Ser}, J., Díaz-Rodríguez, N., and Herrera, F. (2023).
\newblock Explainable artificial intelligence (xai): What we know and what is left to attain trustworthy artificial intelligence.
\newblock {\em Information Fusion}, 99:101805.

\bibitem[Augasta and Kathirvalavakumar, 2011]{rxren}
Augasta, M. and Kathirvalavakumar, T. (2011).
\newblock Reverse engineering the neural networks for rule extraction in classification problems.
\newblock {\em Neural Processing Letters}.

\bibitem[Breiman, 2001]{RF}
Breiman, L. (2001).
\newblock Random forests.
\newblock {\em Mach. Learn.}, 45(1):5–32.

\bibitem[Breiman et~al., 1984]{CART}
Breiman, L., Friedman, J., Olshen, R., and J., S.~C. (1984).
\newblock {\em Classification and Regression Trees}.
\newblock 1st Edition, Chapman and Hall/CRC.

\bibitem[Correa, 2015]{Correa_CSDT}
Correa, B.~A. (2015).
\newblock Example-dependent cost-sensitive decision trees.
\newblock {\em Expert Systems with Applications}, 42:6609--6619.

\bibitem[Craven and Shavlik, 1994]{trepan}
Craven, M.~W. and Shavlik, J.~W. (1994).
\newblock Using sampling and queries to extract rules from trained neural networks.
\newblock In Cohen, W.~W. and Hirsh, H., editors, {\em Machine Learning Proceedings 1994}, pages 37--45. Morgan Kaufmann, San Francisco (CA).

\bibitem[Cubero, 2007]{DTw}
Cubero, J. (2007).
\newblock Weighted classiﬁcation using decision trees for binary classiﬁcation problems.

\bibitem[Dablain et~al., 2024]{Dablain2023}
Dablain, D., Bellinger, C., Krawczyk, B., Aha, W., and Chawla, N. (2024).
\newblock Understanding imbalanced data: Xai and interpretable ml framework.
\newblock {\em Machine Learning}, 113:1--19.

\bibitem[Dragoni et~al., 2020]{dragoni_2020}
Dragoni, M., Donadello, I., and Eccher, C. (2020).
\newblock Explainable ai meets persuasiveness: Translating reasoning results into behavioral change advice.
\newblock {\em Artificial Intelligence in Medicine}, 105:101840.

\bibitem[Elkan, 2001]{Foundations_CSL}
Elkan, C.~P. (2001).
\newblock The foundations of cost-sensitive learning.
\newblock In {\em International Joint Conference on Artificial Intelligence}.

\bibitem[Feng, 1507]{Feng2015}
Feng, S. (2015/07).
\newblock A cost-sensitive decision tree under the condition of multiple classes.
\newblock In {\em Proceedings of the International Conference on Logistics, Engineering, Management and Computer Science}, pages 1212--1218. Atlantis Press.

\bibitem[Freitas, 2014]{freitas}
Freitas, A.~A. (2014).
\newblock Comprehensible classification models: a position paper.
\newblock {\em SIGKDD Explor. Newsl.}, 15(1):1–10.

\bibitem[Friedman, 1937]{Friedman1}
Friedman, M. (1937).
\newblock The use of ranks to avoid the assumption of normality implicit in the analysis of variance.
\newblock {\em Journal of the American Statistical Association}, 32(200):675--701.

\bibitem[Friedman, 1940]{Friedman2}
Friedman, M. (1940).
\newblock {A Comparison of Alternative Tests of Significance for the Problem of $m$ Rankings}.
\newblock {\em The Annals of Mathematical Statistics}, 11(1):86 -- 92.

\bibitem[Guidotti et~al., 2018]{Guidotti2019}
Guidotti, R., Monreale, A., Ruggieri, S., Turini, F., Giannotti, F., and Pedreschi, D. (2018).
\newblock A survey of methods for explaining black box models.
\newblock {\em ACM Comput. Surv.}, 51(5).

\bibitem[Gunning and Aha, 2019]{darpa}
Gunning, D. and Aha, D. (2019).
\newblock Darpa’s explainable artificial intelligence (xai) program.
\newblock {\em AI Magazine}, 40(2):44--58.

\bibitem[Haomin~Wang and Peng, 2021]{Wang2021}
Haomin~Wang, G.~K. and Peng, Y. (2021).
\newblock Multi-class misclassification cost matrix for credit ratings in peer-to-peer lending.
\newblock {\em Journal of the Operational Research Society}, 72(4):923--934.

\bibitem[Huysmans et~al., 2011]{Huysmans_2011}
Huysmans, J., Dejaeger, K., Mues, C., Vanthienen, J., and Baesens, B. (2011).
\newblock An empirical evaluation of the comprehensibility of decision table, tree and rule based predictive models.
\newblock {\em Decision Support Systems}, 51(1):141--154.

\bibitem[Kopanja, 2024]{kopanja2024}
Kopanja, M. (2024).
\newblock A novel model-agnostic xai method guided by cost-sensitive tree models and argumentative decision graphs.
\newblock In {\em Joint Proceedings of the xAI 2024 Late-breaking Work, Demos and Doctoral Consortium co-located with the 2nd World Conference on eXplainable Artificial Intelligence (xAI 2024)}, pages 393--400.

\bibitem[Kopanja et~al., 2023]{kopanja2023}
Kopanja, M., Brdar, S., and Hacko, S. (2023).
\newblock Uncovering decision-making process of cost-sensitive tree-based classifiers using the adaptation of treeshap.
\newblock In {\em "Joint Proceedings of the xAI-2023 Late-breaking Work, Demos and Doctoral Consortium co-located with the 1st World Conference on eXplainable Artificial Intelligence (xAI-2023)"}, pages 95--100.

\bibitem[Kopanja et~al., 2024]{kopanja}
Kopanja, M., Hačko, S., Brdar, S., and Savić, M. (2024).
\newblock Cost-sensitive tree shap for explaining cost-sensitive tree-based models.
\newblock {\em Computational Intelligence}, 40(3):e12651.

\bibitem[Krawczyk et~al., 2014]{Krawczyk2014}
Krawczyk, B., Woźniak, M., and Schaefer, G. (2014).
\newblock Cost-sensitive decision tree ensembles for effective imbalanced classification.
\newblock {\em Applied Soft Computing}, 14:554--562.

\bibitem[Lomax and Vadera, 2011]{Lomax_2011}
Lomax, S. and Vadera, S. (2011).
\newblock An empirical comparison of cost-sensitive decision tree induction algorithms.
\newblock {\em Expert Systems}, 28:227--268.

\bibitem[Longo et~al., 2024]{manifesto}
Longo, L., Brcic, M., Cabitza, F., Choi, J., Confalonieri, R., Ser, J.~D., Guidotti, R., Hayashi, Y., Herrera, F., Holzinger, A., Jiang, R., Khosravi, H., Lecue, F., Malgieri, G., Páez, A., Samek, W., Schneider, J., Speith, T., and Stumpf, S. (2024).
\newblock Explainable artificial intelligence (xai) 2.0: A manifesto of open challenges and interdisciplinary research directions.
\newblock {\em Information Fusion}, 106:102301.

\bibitem[Longo et~al., 2020]{luca_review}
Longo, L., Goebel, R., Lecue, F., Kieseberg, P., and Holzinger, A. (2020).
\newblock Explainable artificial intelligence: Concepts, applications, research challenges and visions.
\newblock In Holzinger, A., Kieseberg, P., Tjoa, A.~M., and Weippl, E., editors, {\em Machine Learning and Knowledge Extraction}, pages 1--16, Cham. Springer International Publishing.

\bibitem[Lopes et~al., 2022]{Lopes_2022}
Lopes, P., Silva, E., Braga, C., Oliveira, T., and Rosado, L. (2022).
\newblock Xai systems evaluation: A review of human and computer-centred methods.
\newblock {\em Applied Sciences}, 12(19).

\bibitem[Lundberg et~al., 2020]{TreeSHAP}
Lundberg, S., Erion, G., Chen, H., DeGrave, A., Prutkin, J., Nair, B., Katz, R., Himmelfarb, J., Bansal, N., and Lee, S. (2020).
\newblock From local explanations to global understanding with explainable ai for trees.
\newblock {\em Nature Machine Intelligence}, 2(1):2522--5839.

\bibitem[Lundberg and Lee, 2017]{shap}
Lundberg, S. and Lee, S. (2017).
\newblock A unified approach to interpreting model predictions.
\newblock In {\em Advances in Neural Information Processing Systems 30}, pages 4765--4774. Curran Associates, Inc.

\bibitem[Maouche et~al., 2023]{Maouche2023}
Maouche, I., Terrissa, L.~S., Benmohammed, K., and Zerhouni, N. (2023).
\newblock An explainable ai approach for breast cancer metastasis prediction based on clinicopathological data.
\newblock {\em IEEE Transactions on Biomedical Engineering}, 70(12):3321--3329.

\bibitem[Martens et~al., 2007]{svm_rules}
Martens, D., Baesens, B., {Van Gestel}, T., and Vanthienen, J. (2007).
\newblock Comprehensible credit scoring models using rule extraction from support vector machines.
\newblock {\em European Journal of Operational Research}, 183(3):1466--1476.

\bibitem[Mekonnen et~al., 2023]{DT_surrogate}
Mekonnen, E., Dondio, P., and Longo, L. (2023).
\newblock Explaining deep learning time series classification models using a decision tree-based post-hoc xai method.
\newblock volume 3554. CEUR-WS.
\newblock Publisher Copyright: {\textcopyright} 2023 CEUR-WS. All rights reserved.; Joint 1st World Conference on eXplainable Artificial Intelligence: Late-Breaking Work, Demos and Doctoral Consortium, xAI-2023: LB-D-DC ; Conference date: 26-07-2023 Through 28-07-2023.

\bibitem[Mustari et~al., 2023]{Mustari2023}
Mustari, A., Ahmed, R., Tasnim, A., Juthi, J.~S., and Shahariar, G.~M. (2023).
\newblock Explainable contrastive and cost-sensitive learning for cervical cancer classification.
\newblock In {\em 2023 26th International Conference on Computer and Information Technology (ICCIT)}, pages 1--6.

\bibitem[Parimbelli et~al., 2023]{Araucana}
Parimbelli, E., Buonocore, T.~M., Nicora, G., Michalowski, W., Wilk, S., and Bellazzi, R. (2023).
\newblock Why did ai get this one wrong? — tree-based explanations of machine learning model predictions.
\newblock {\em Artificial Intelligence in Medicine}, 135:102471.

\bibitem[Pedregosa et~al., 2011]{scikit-learn}
Pedregosa, F., Varoquaux, G., Gramfort, A., Michel, V., Thirion, B., Grisel, O., Blondel, M., Prettenhofer, P., Weiss, R., Dubourg, V., Vanderplas, J., Passos, A., Cournapeau, D., Brucher, M., Perrot, M., and Duchesnay, E. (2011).
\newblock Scikit-learn: Machine learning in {P}ython.
\newblock {\em Journal of Machine Learning Research}, 12:2825--2830.

\bibitem[Qin et~al., 2005]{Qin2005}
Qin, Z., Zhang, S., and Zhang, C. (2005).
\newblock Cost-sensitive decision trees with multiple cost scales.
\newblock In Webb, G.~I. and Yu, X., editors, {\em AI 2004: Advances in Artificial Intelligence}, pages 380--390, Berlin, Heidelberg. Springer Berlin Heidelberg.

\bibitem[Ranjbar and Safabakhsh, 2022]{tree_alime}
Ranjbar, N. and Safabakhsh, R. (2022).
\newblock Using decision tree as local interpretable model in autoencoder-based lime.

\bibitem[Ribeiro et~al., 2016a]{ribeiro2016}
Ribeiro, M.~T., Singh, S., and Guestrin, C. (2016a).
\newblock Model-agnostic interpretability of machine learning.

\bibitem[Ribeiro et~al., 2016b]{lime}
Ribeiro, M.~T., Singh, S., and Guestrin, C. (2016b).
\newblock "why should i trust you?": Explaining the predictions of any classifier.
\newblock In {\em Proceedings of the 22nd ACM SIGKDD International Conference on Knowledge Discovery and Data Mining}, KDD '16, page 1135–1144, New York, NY, USA. Association for Computing Machinery.

\bibitem[Ribeiro et~al., 2018]{Anchors}
Ribeiro, M.~T., Singh, S., and Guestrin, C. (2018).
\newblock Anchors: High-precision model-agnostic explanations.
\newblock {\em Proceedings of the AAAI Conference on Artificial Intelligence}, 32(1).

\bibitem[Rivera-López and Ceballos, 2024]{Rivera2024}
Rivera-López, R. and Ceballos, H.~G. (2024).
\newblock A differential-evolution-based approach to extract univariate decision trees from black-box models using tabular data.
\newblock {\em IEEE Access}, 12:169850--169868.

\bibitem[Rizzo and Longo, 2018]{RizzoL18Explainability}
Rizzo, L. and Longo, L. (2018).
\newblock A qualitative investigation of the explainability of defeasible argumentation and non-monotonic fuzzy reasoning.
\newblock In {\em Proceedings for the 26th {AIAI} Irish Conference on Artificial Intelligence and Cognitive Science Trinity College Dublin, Dublin, Ireland, December 6-7th, 2018.}, pages 138--149.

\bibitem[Savić et~al., 2022]{Savic2022}
Savić, M., Atanasijević, J., Jakovetić, D., and Krejić, N. (2022).
\newblock Tax evasion risk management using a hybrid unsupervised outlier detection method.
\newblock {\em Expert Systems with Applications}, 193:116409.

\bibitem[Schmitz et~al., 1999]{Schmitz1999}
Schmitz, G., Aldrich, C., and Gouws, F. (1999).
\newblock Ann-dt: an algorithm for extraction of decision trees from artificial neural networks.
\newblock {\em IEEE Transactions on Neural Networks}, 10(6):1392--1401.

\bibitem[Shankaranarayana and Runje, 2019]{alime}
Shankaranarayana, S.~M. and Runje, D. (2019).
\newblock Alime: Autoencoder based approach for local interpretability.
\newblock In Yin, H., Camacho, D., Tino, P., Tall{\'o}n-Ballesteros, A.~J., Menezes, R., and Allmendinger, R., editors, {\em Intelligent Data Engineering and Automated Learning -- IDEAL 2019}, pages 454--463, Cham. Springer International Publishing.

\bibitem[Sun et~al., 2006]{Sun_2006}
Sun, Y., Kamel, M.~S., and Wang, Y. (2006).
\newblock Boosting for learning multiple classes with imbalanced class distribution.
\newblock In {\em Sixth International Conference on Data Mining (ICDM'06)}, pages 592--602.

\bibitem[Sun et~al., 2011]{imbl_review}
Sun, Y., Wong, A., and Kamel, M.~S. (2011).
\newblock Classification of imbalanced data: a review.
\newblock {\em International Journal of Pattern Recognition and Artificial Intelligence}, 23.

\bibitem[Turney, 1995]{Turney1995}
Turney, P. (1995).
\newblock Cost-sensitive classification: Empirical evaluation of a hybrid genetic decision tree induction algorithm.
\newblock {\em Journal of Artificial Intelligence Research}, 2:369--409.

\bibitem[Turney, 2002]{Turney}
Turney, P.~D. (2002).
\newblock Types of cost in inductive concept learning.

\bibitem[Vilone and Longo, 2020]{vilone_review}
Vilone, G. and Longo, L. (2020).
\newblock Explainable artificial intelligence: a systematic review.

\bibitem[Vilone and Longo, 2021a]{Vilone2021}
Vilone, G. and Longo, L. (2021a).
\newblock Classification of explainable artificial intelligence methods through their output formats.
\newblock {\em Machine Learning and Knowledge Extraction}, 3(3):615--661.

\bibitem[Vilone and Longo, 2021b]{giulia_2021}
Vilone, G. and Longo, L. (2021b).
\newblock A quantitative evaluation of global, rule-based explanations of post-hoc, model agnostic methods.
\newblock {\em Frontiers in Artificial Intelligence}, 4.

\bibitem[Vilone and Longo, 2023]{giulia_test}
Vilone, G. and Longo, L. (2023).
\newblock Development of a human-centred psychometric test for the evaluation of explanations produced by xai methods.
\newblock In Longo, L., editor, {\em Explainable Artificial Intelligence - 1st World Conference, xAI 2023, Proceedings}, Communications in Computer and Information Science, pages 205--232, Germany. Springer Science and Business Media Deutschland GmbH.

\bibitem[Vilone et~al., 2020]{giulia_2020}
Vilone, G., Rizzo, L., and Longo, L. (2020).
\newblock A comparative analysis of rule-based, model-agnostic methods for explainable artificial intelligence.
\newblock {\em Proceedings for the 28th AIAI Irish Conference on Artificial Intelligence and Cognitive Science, Dublin, Ireland}, 2771.

\bibitem[Wilcoxon, 1945]{Wilcoxon}
Wilcoxon, F. (1945).
\newblock Individual comparisons by ranking methods.
\newblock {\em Biometrics Bulletin}, 1(6):80--83.

\bibitem[Zhou and Jiang, 2003]{c45}
Zhou, Z.-H. and Jiang, Y. (2003).
\newblock Medical diagnosis with c4.5 rule preceded by artificial neural network ensemble.
\newblock {\em IEEE transactions on information technology in biomedicine : a publication of the IEEE Engineering in Medicine and Biology Society}, 7:37--42.

\bibitem[Zhou et~al., 2003]{refne}
Zhou, Z.-H., Jiang, Y., and Chen, S.-F. (2003).
\newblock Extracting symbolic rules from trained neural network ensembles.
\newblock {\em AI Commun.}, 16(1):3–15.

\end{thebibliography}

\newpage
\onecolumn
\appendix
\section{Quantitative measures for six rule-extractors generated for NN-2 models across datasets.}

\begin{figure*}[htb!]
    \centering
    \includegraphics[width=0.8\linewidth]{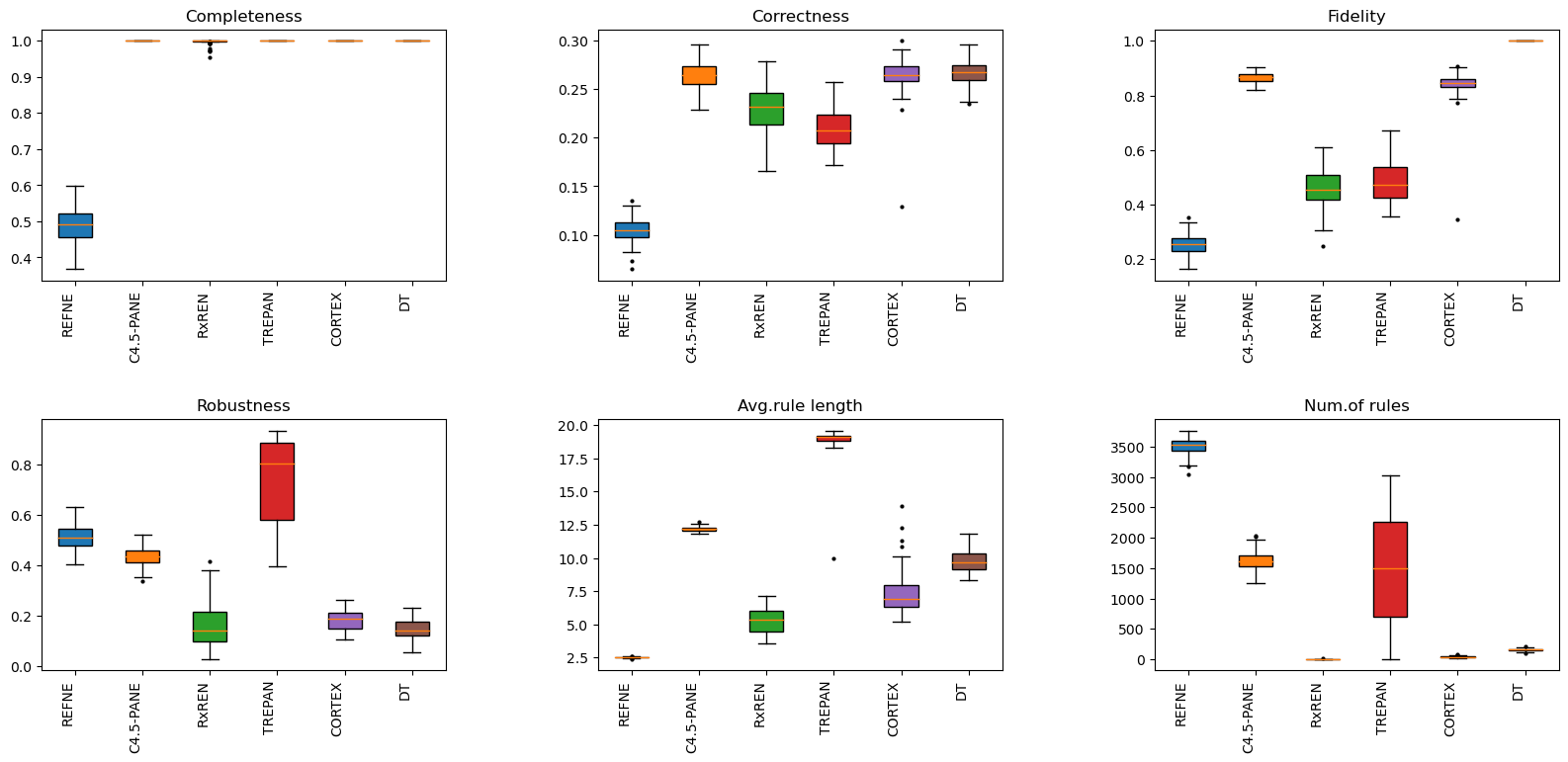}
    \caption{Distribution of six quantitative metrics for evaluating rule-extractor XAI methods for abalone dataset.}
    \label{all_metrics_datasets}
\end{figure*}

\begin{figure*}[htb!]
    \centering
    \includegraphics[width=0.8\linewidth]{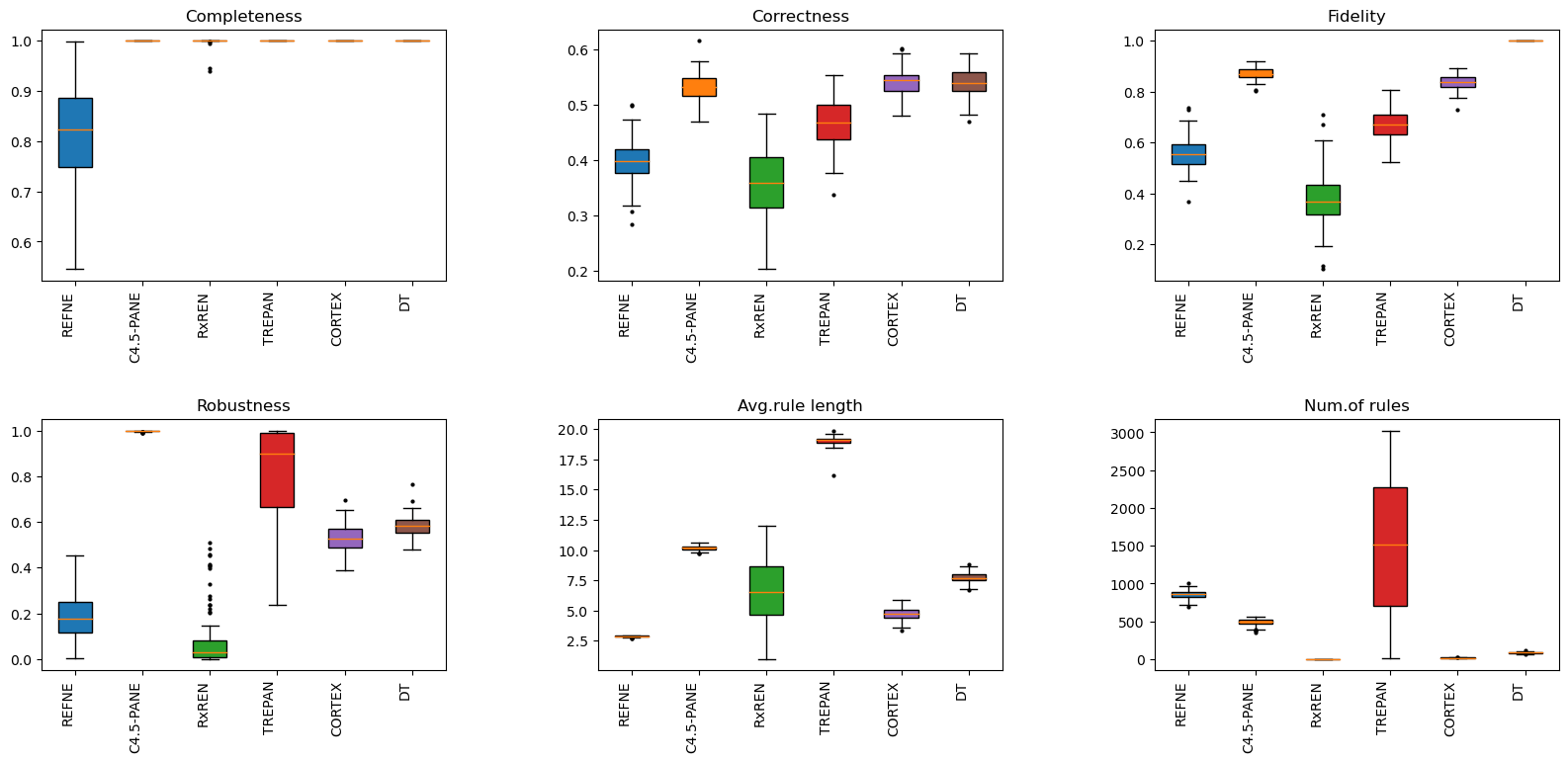}
    \caption{Distribution of six quantitative metrics for evaluating rule-extractor XAI methods for the contraceptive dataset.}
    \label{all_metrics_datasets}
\end{figure*}

\begin{figure*}[htb!]
    \centering
    \includegraphics[width=0.8\linewidth]{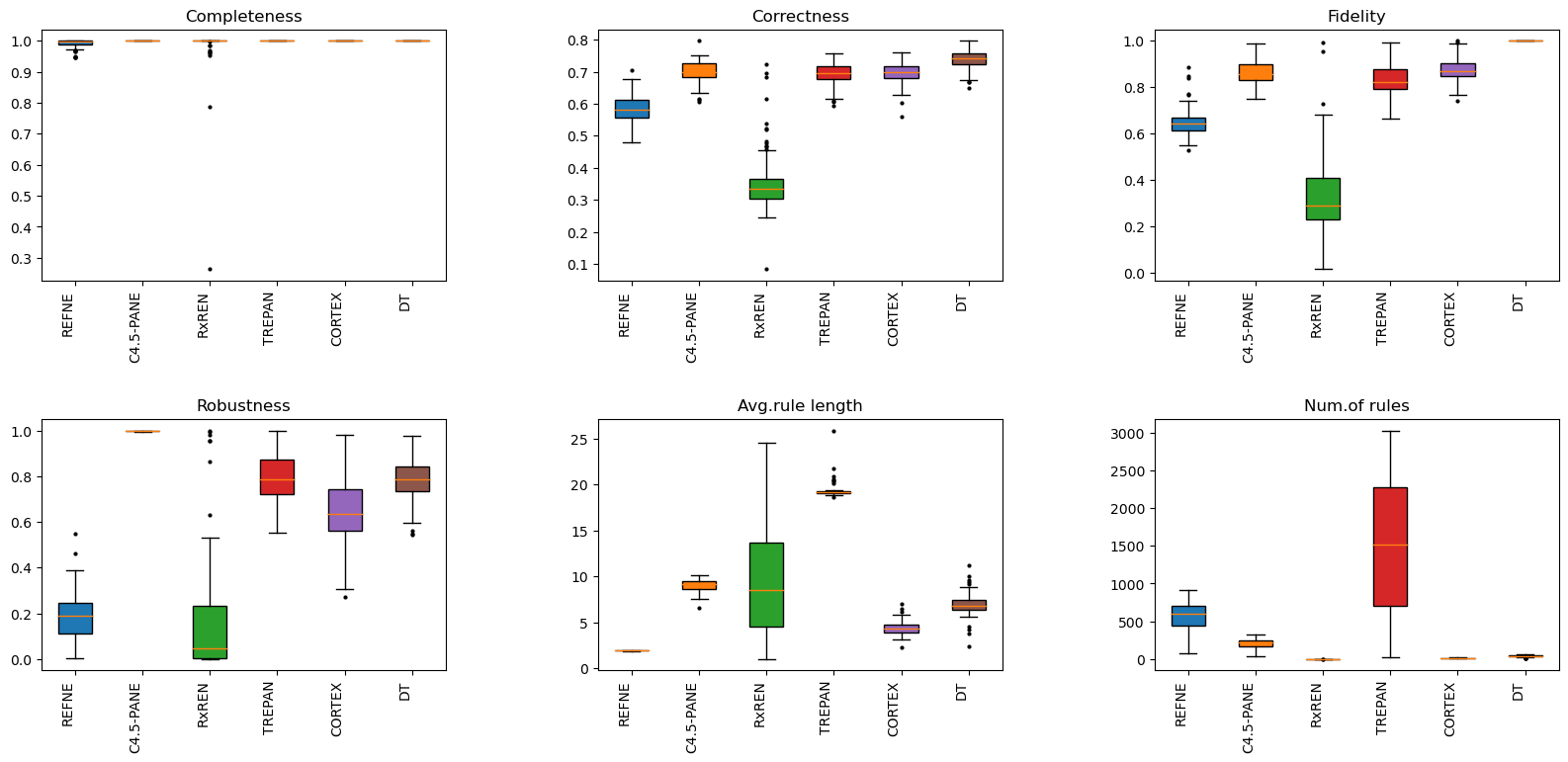}
    \caption{Distribution of six quantitative metrics for evaluating rule-extractor XAI methods for credit dataset.}
    \label{all_metrics_datasets}
\end{figure*}

\begin{figure*}[htb!]
    \centering
    \includegraphics[width=0.8\linewidth]{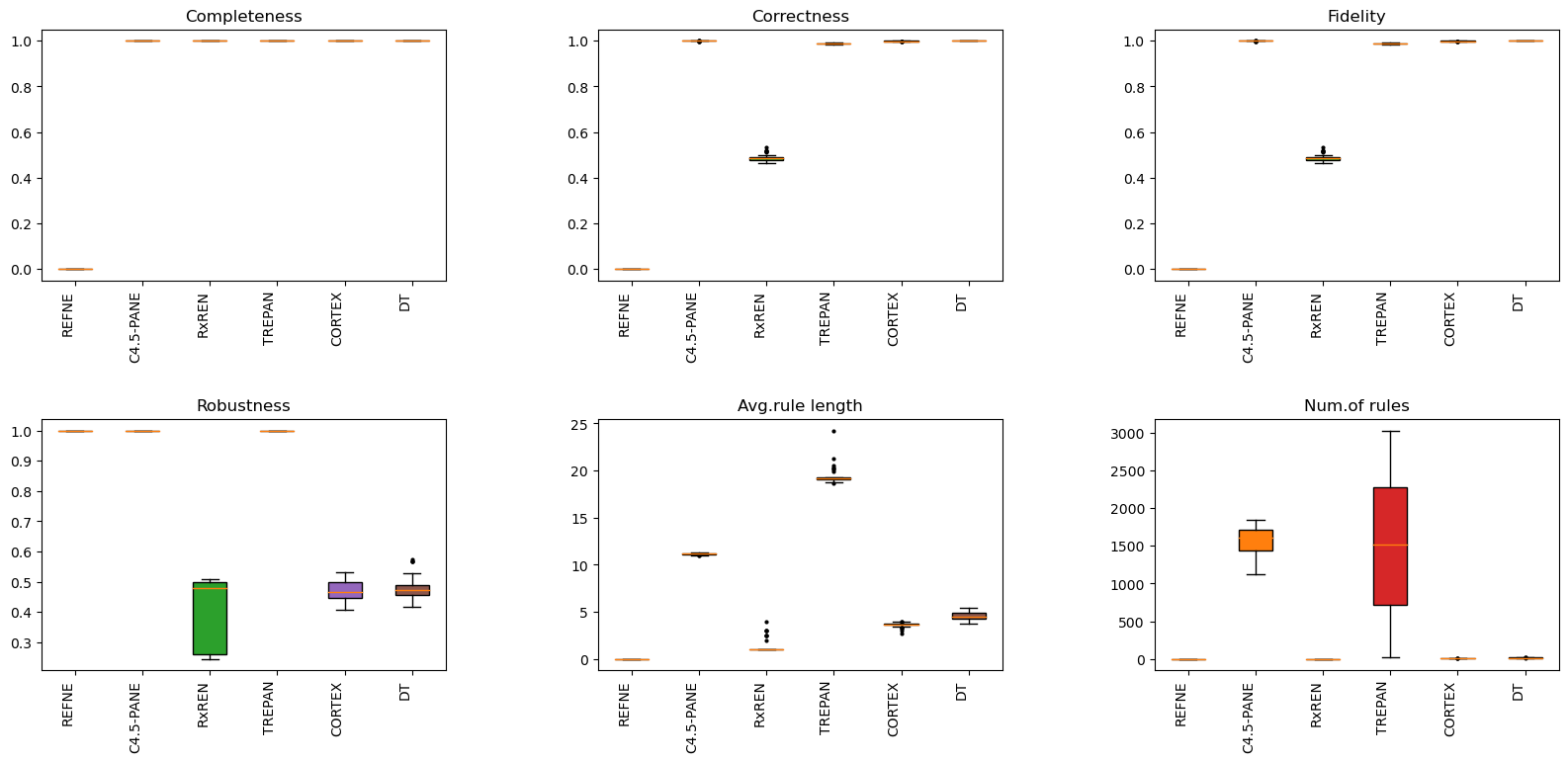}
    \caption{Distribution of six quantitative metrics for evaluating rule-extractor XAI methods for the mushroom dataset.}
    \label{all_metrics_datasets}
\end{figure*}

\begin{figure*}[htb!]
    \centering
    \includegraphics[width=0.8\linewidth]{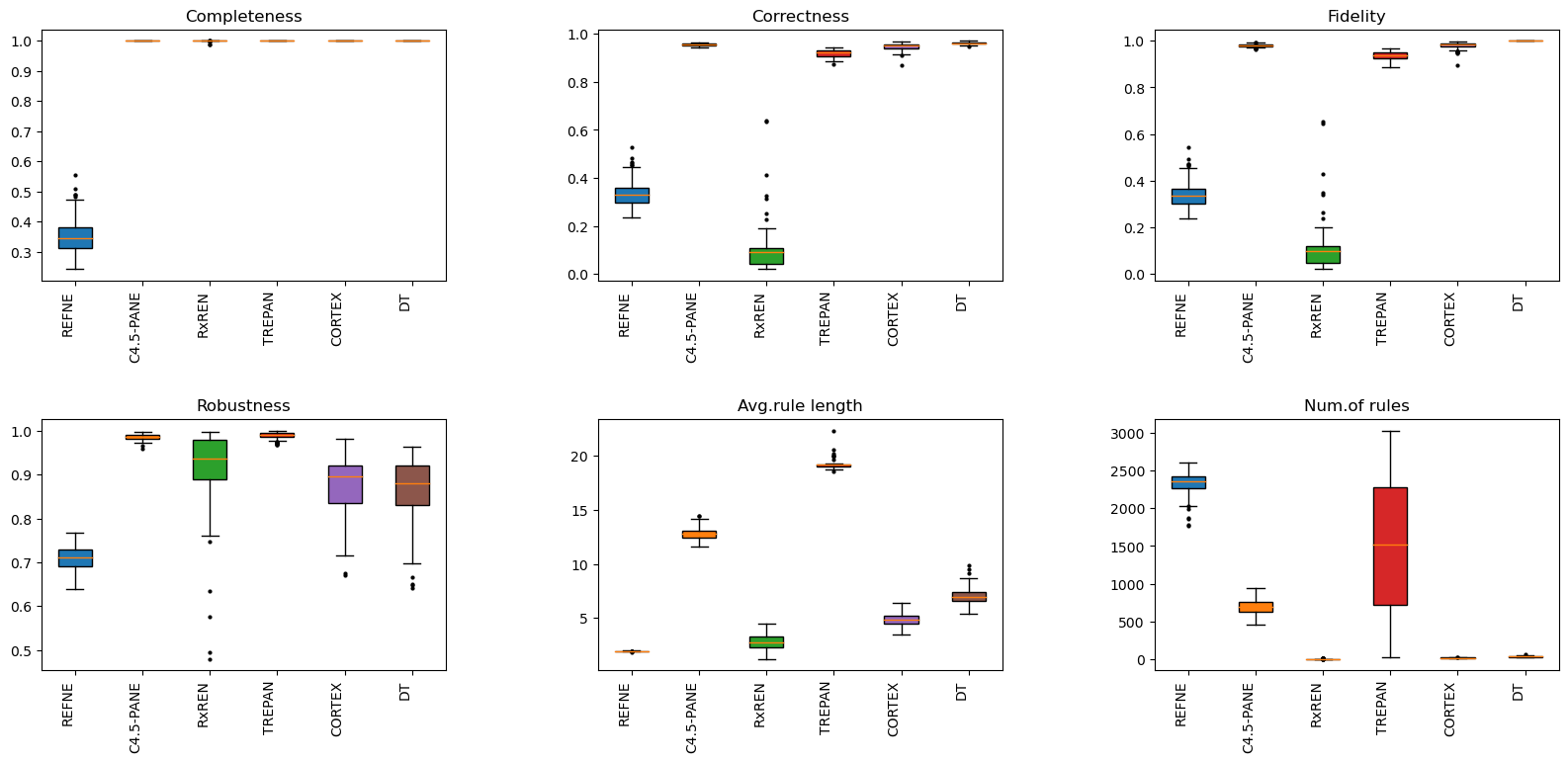}
    \caption{Distribution of six quantitative metrics for evaluating rule-extractor XAI methods for page-blocks dataset.}
    \label{all_metrics_datasets}
\end{figure*}

\begin{figure*}[htb!]
    \centering
    \includegraphics[width=0.8\linewidth]{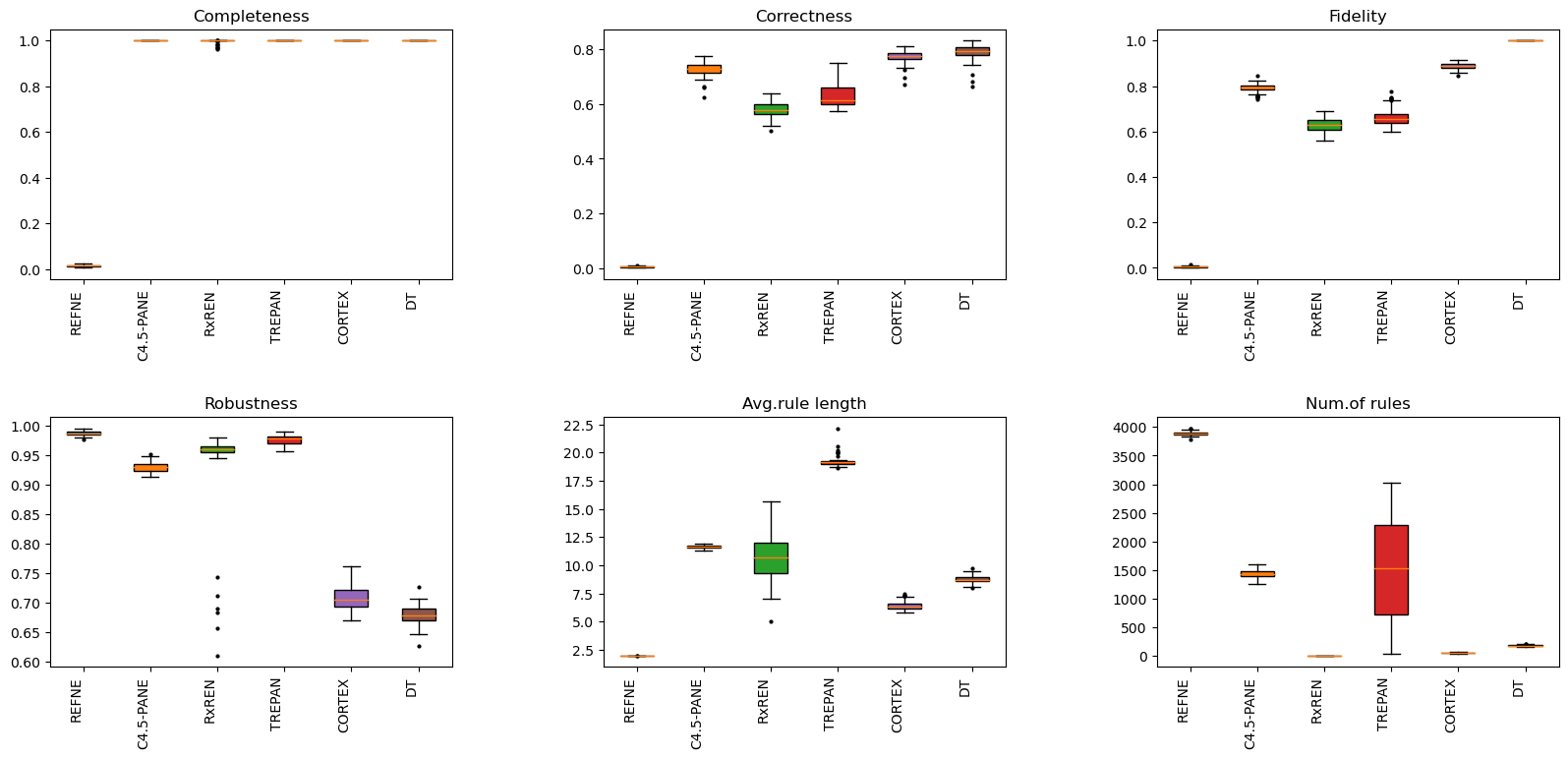}
    \caption{Distribution of six quantitative metrics for evaluating rule-extractor XAI methods for the wave-form dataset.}
    \label{all_metrics_datasets}
\end{figure*}

\begin{figure*}[htb!]
    \centering
    \includegraphics[width=0.8\linewidth]{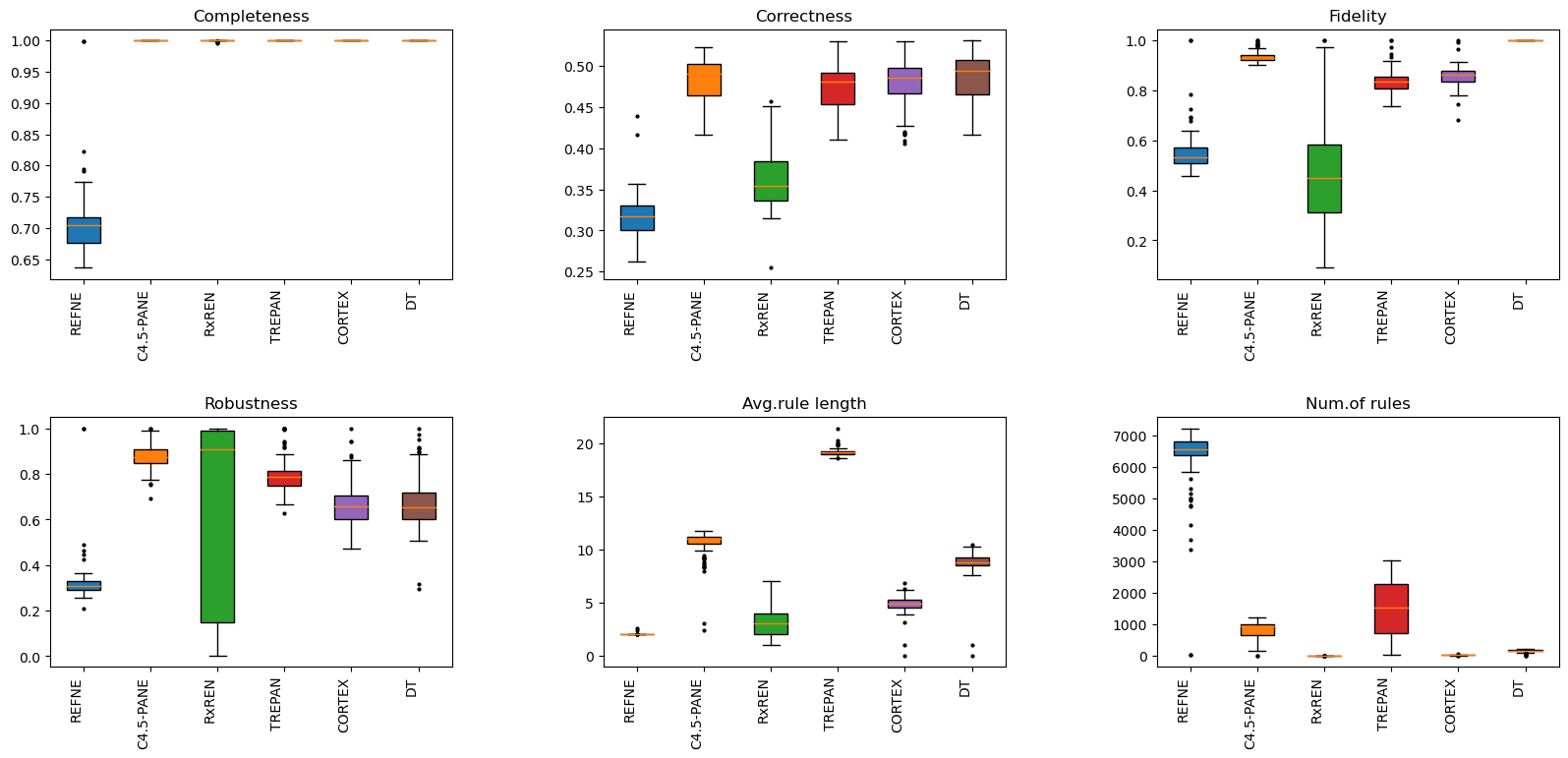}
    \caption{Distribution of six quantitative metrics for evaluating rule-extractor XAI methods for wine dataset.}
    \label{all_metrics_datasets}
\end{figure*}

\begin{figure*}[htb!]
    \centering
    \includegraphics[width=0.8\linewidth]{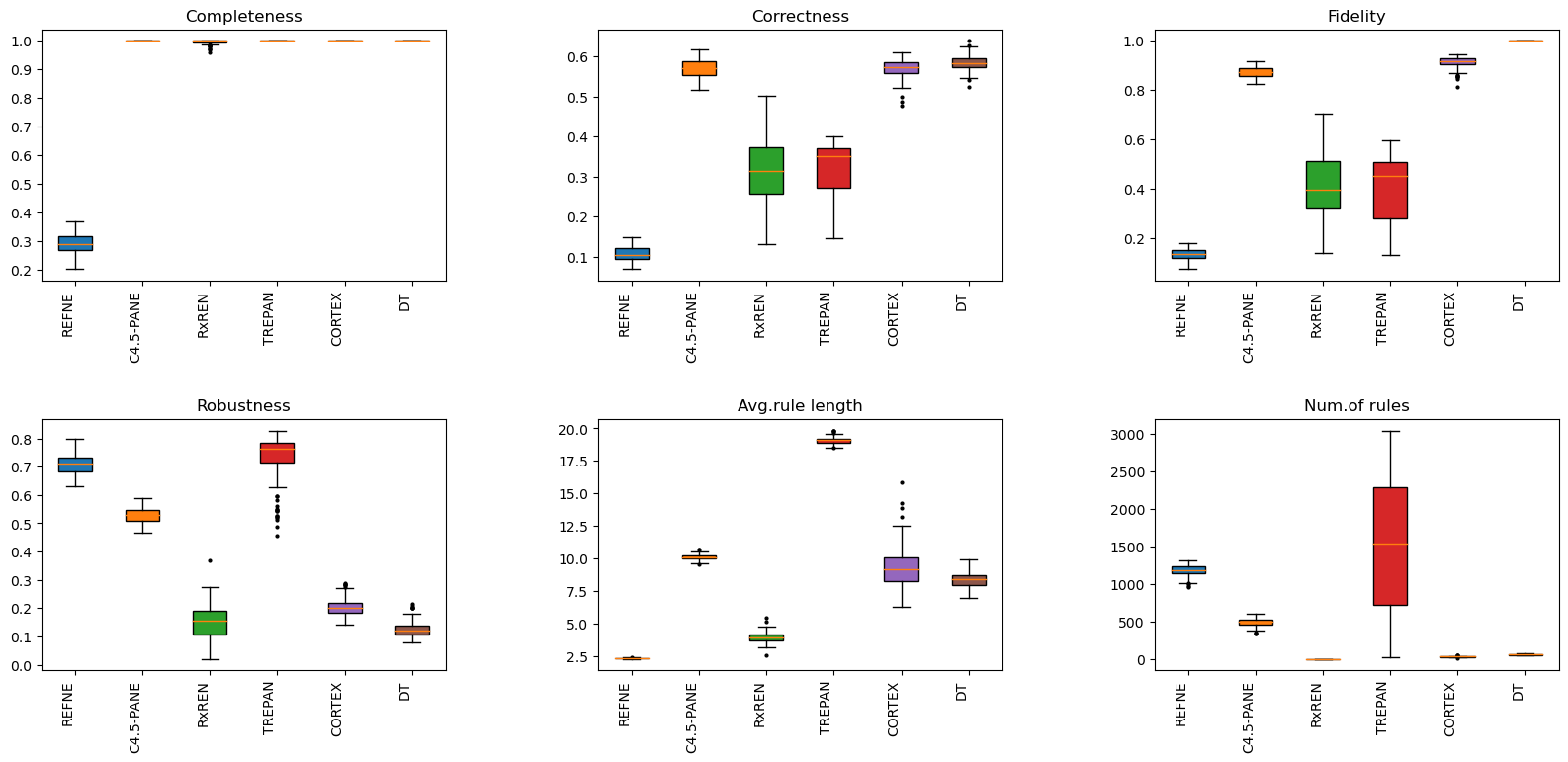}
    \caption{Distribution of six quantitative metrics for evaluating rule-extractor XAI methods for yeast dataset.}
    \label{all_metrics_datasets}
\end{figure*}

\newpage
\onecolumn
\section{Example of a set of rules generated by CORTEX method.}
\begin{figure*}[h!]
    \centering
    \includegraphics[width=0.7\linewidth]{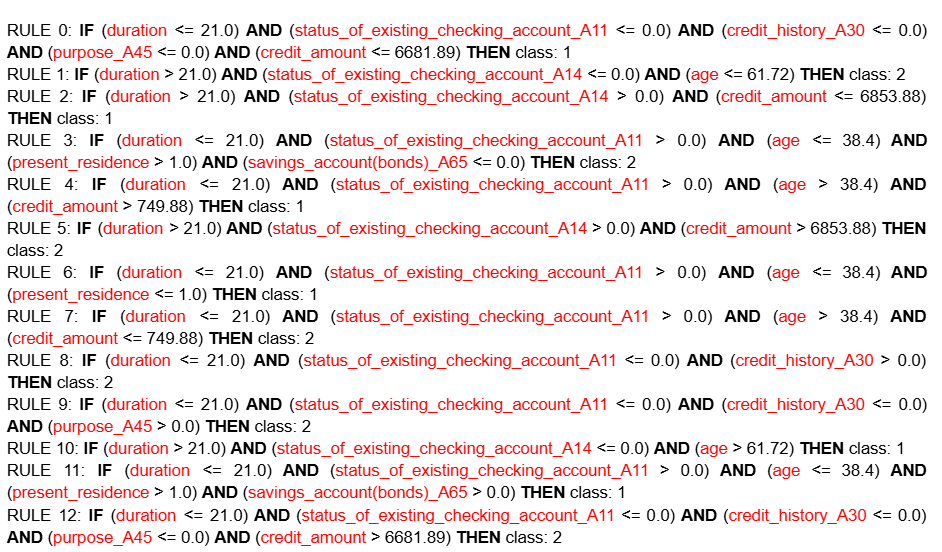}
    \caption{Rule set extracted by CORTEX method on credit dataset.}
    \label{rules}
\end{figure*}

\end{document}